\documentclass[10pt,journal,final]{IEEEtran}

\ifCLASSINFOpdf

\usepackage[pdftex]{graphicx}
\usepackage{color}
\usepackage{multirow}
\usepackage{booktabs}
\usepackage{enumerate}
\usepackage{array}
\usepackage{float}

\else
\usepackage[utf8]{inputenc}

\fi
\usepackage{amsfonts,amssymb}
\usepackage{amsmath}
\usepackage{algorithm}
\usepackage{algorithmicx}

\ifCLASSOPTIONcompsoc
\usepackage[caption=false,font=normalsize,labelfont=sf,textfont=sf]{subfig}
\else
\usepackage[caption=false,font=footnotesize]{subfig}
\fi
\usepackage{cite}
\usepackage{dblfloatfix}
\begin{document}
	
	\title{Remote Sensing Object Detection Meets Deep Learning: A Meta-review of Challenges and Advances}
	
	%
	%
	%
	
	\author{
			Xiangrong~Zhang,~\IEEEmembership{Senior~Member,~IEEE},
			Tianyang~Zhang,
			Guanchun~Wang,
			Peng~Zhu,
			Xu~Tang,~\IEEEmembership{Senior~Member,~IEEE},
			Xiuping~Jia,~\IEEEmembership{Fellow,~IEEE}, 
			and~Licheng~Jiao,~\IEEEmembership{Fellow,~IEEE} 
		\thanks{Xiangrong Zhang, Tianyang Zhang, Guanchun Wang, Peng Zhu, Xu Tang, and Licheng Jiao are with the School of Artificial Intelligence, Xidian University, Xi’an 710071, China (e-mail: xrzhang@mail.xidian.edu.cn).
			
		Xiuping Jia is with the School of Engineering and Information Technology, University of New South Wales, Canberra, ACT 2612, Australia.}
		}%

	%



	\maketitle
	
	\begin{abstract}		
		Remote sensing object detection (RSOD), one of the most fundamental and challenging tasks in the remote sensing field, has received longstanding attention. In recent years, deep learning techniques have demonstrated robust feature representation capabilities and led to a big leap in the development of RSOD techniques. In this era of rapid technical evolution, this review aims to present a comprehensive review of the recent achievements in deep learning based RSOD methods. More than 300 papers are covered in this review. We identify five main challenges in RSOD, including multi-scale object detection, rotated object detection, weak object detection, tiny object detection, and object detection with limited supervision, and systematically review the corresponding methods developed in a hierarchical division manner.  We also review the widely used benchmark datasets and evaluation metrics within the field of RSOD, as well as the application scenarios for RSOD.
		Future research directions are provided for further promoting the research in RSOD.		
	\end{abstract}
	
	\begin{IEEEkeywords}
		Object detection,
		Remote sensing images,
		Deep learning,
		Technical evolution
	\end{IEEEkeywords}
	
	\begin{figure*}[tb!]
		\begin{center}
		\includegraphics[width=0.98\textwidth]{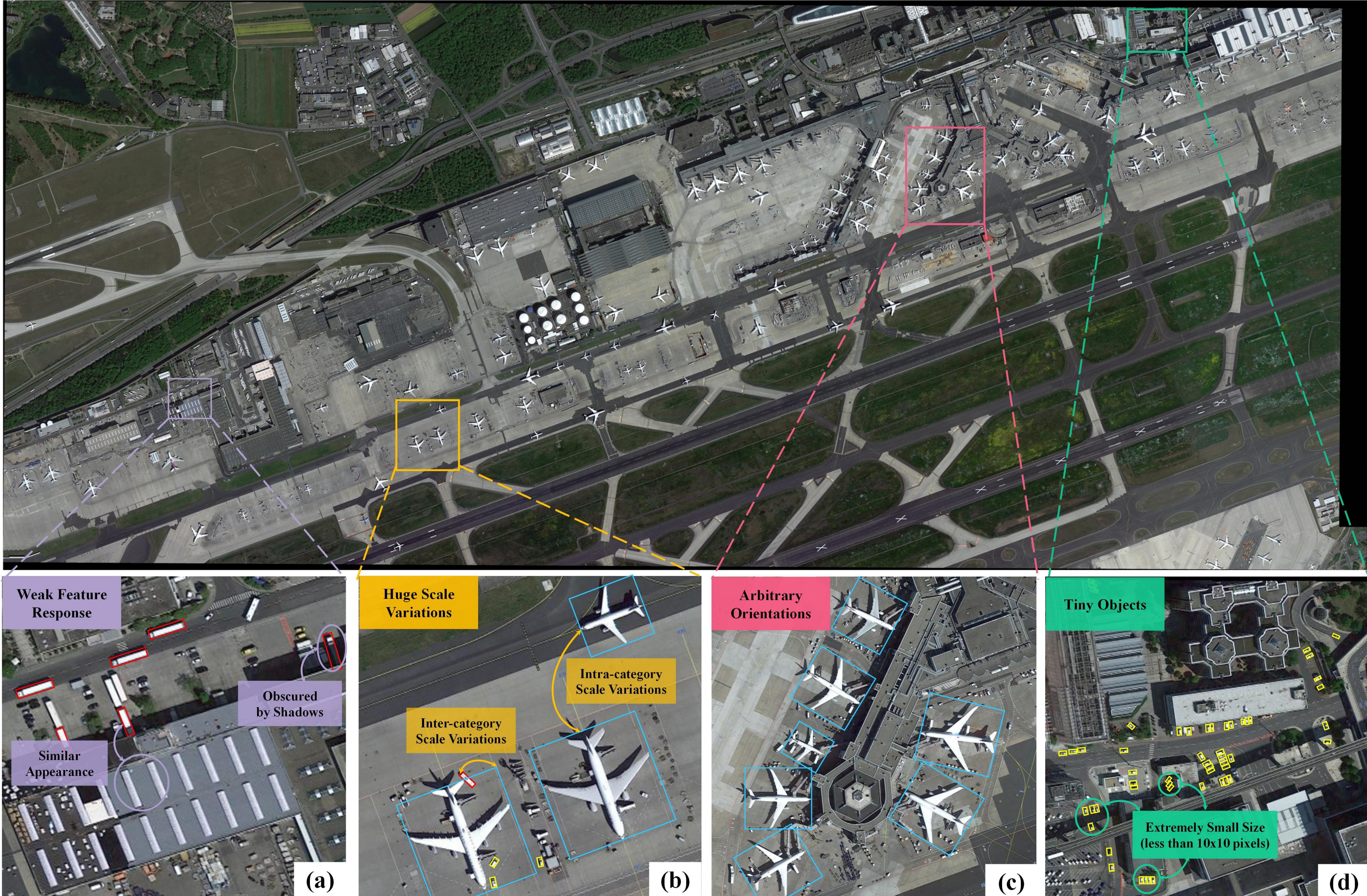}
		\end{center}
		\vspace{-3mm}
		\caption{A typical example of remote sensing images. 
		(a) The complex context and massive background noises lead to weak feature responses of objects.
		(b) Huge scale variations exist in both inter-category and intra-category objects.
		(c) Objects are distributed with arbitrary orientations. 
		(d) Tiny objects tend to exhibit extremely small scales.}
		\label{fig:introduction}
	\end{figure*}
	\section{Introduction}

	With the rapid advances in earth observation technology, remote sensing satellites (e.g., Google Earth \cite{google}, WordWide-3 \cite{xView}, and Gaofen series satellites \cite{Gaofen1,Gaofen2,Gaofen3}) have made significant improvements in spatial, temporal, and spectral resolutions and a massive number of remote sensing images (RSIs) are now accessible. Benefiting from the dramatic increase in available RSIs, human beings have entered an era of remote sensing big data, and the automatic interpretation of RSIs has become an active yield challenging topic \cite{zhu2017deep, DLforRS, AIforRS}.
	
	RSOD aims to determine whether or not objects of interest exist in a given RSI and to return the category and position of each predicted object. The term 'object' in this survey refers to man-made or highly structured objects (such as airplanes, vehicles, and ships) rather than unstructured scene objects (e.g., land, sky, and grass). As the cornerstone in the automatic interpretation of RSIs, RSOD has received significant attention.
	
	In general, RSIs are taken at an overhead viewpoint with different ground sampling distances (GSDs) and cover widespread regions of the Earth's surface. As a result, the geospatial objects exhibit more significant diversity in scale, angle, and appearance. Based on the characteristics of geospatial objects in RSIs, we summarize the major challenges of RSOD in the following five aspects:

	(1) \textbf{Huge Scale Variations.} On the one hand, there are generally massive scale variations across different categories of objects, as illustrated in Fig. \ref{fig:introduction}(b): a vehicle may be as small as 10 pixel area, while an airplane can be 20 times larger than the vehicle. On the other hand, the intra-category objects also show a wide range of scales. Therefore, the detection models require to handle both large-scale and small-scale objects.
		
	(2) \textbf{Arbitrary Orientations.} The unique overhead viewpoint leads to the geospatial objects often distributed with arbitrary orientations, as shown in Fig. \ref{fig:introduction}(c). This rotated object detection task exacerbates the challenge of RSOD, making it important for the detector to be perceptive of orientation.

	(3) \textbf{Weak Feature Responses.} Generally, RSIs contain complex context and massive background noises. As depicted in Fig. \ref{fig:introduction}(a), some vehicles are obscured by shadows, and the surrounding background noises tend to have a similar appearance to vehicles. This intricate interference may overwhelm the objects of interest and deteriorate their feature representation, which results in the objects of interest being presented as weak feature responses \cite{weak-reponse-survey}.

	(4) \textbf{Tiny Objects.} As shown in Fig. \ref{fig:introduction}(d), tiny objects tend to exhibit extremely small scales and limited appearance information, resulting in a poor-quality feature representation. In addition, the current prevailing detection paradigms inevitably weaken or even discard the representation of tiny objects \cite{NWD}. These problems in tiny object detection bring new difficulties to existing detection methods. 
	
	(5) \textbf{Expensive Annotation.} The complex characteristics of geospatial objects in terms of scale and angle, as well as the expert knowledge required for fine-grained annotations \cite{weak-supervision-survey}, make the accurate box-level annotations of RSIs a time-consuming and labor-intensive task. However, the current deep learning based detectors rely heavily on abundant well-labeled data to reach performance saturation. Therefore, the efficient RSOD methods in a lack of sufficient supervised information scenario remain challenging.
	
	To tackle these challenges, numerous RSOD methods have emerged in the past two decades. At the early stage, researchers adopted template matching \cite{Template1, Template2, Template3} and prior knowledge \cite{Knowledge1, Knowledge2, Knowledge3} for object detection in remote sensing scenes. These early methods rely more on hand-crafted templates or prior knowledge, leading to unstable results. Later, machine learning approaches \cite{ML-1, ML-2, ML-3, ML-4} have become mainstream in RSOD, which view object detection as a classification task. Concretely, the machine learning model first searches a set of object proposals from the input image and extracts the texture, context, and other features of these object proposals. Then, it employs an independent classifier to identify the object categories in these object proposals. However, shallow learning based features from the machine learning approaches significantly restrict the representations of objects, especially in more challenging scenarios. Besides, the machine learning based object detection methods cannot be trained in an end-to-end manner, which is no longer applicable in the era of remote sensing big data.

	Recently, deep learning techniques \cite{DeepLearning} have demonstrated powerful feature representation capabilities from massive amounts of data, and the state-of-the-art detectors \cite{Faster-RCNN,YOLO,focal-loss, FCOS} in computer vision achieve object detection ability that rivals that of humans \cite{OD-Survey-Ouyangwanli}. Drawing on the advanced progress of deep learning techniques, various deep learning based methods have dominated the RSOD and led to remarkable breakthroughs in detection performance. Compared to the traditional methods, deep neural network architecture can extract high-level semantic features and obtain much more robust feature representations of objects. In addition, the efficient end-to-end training manner and automated feature extraction fashion make the deep learning based object detection methods more suitable for RSOD in the remote sensing big data era.
	
	\begin{figure*}[ht]
		\begin{center}
		\includegraphics[width=0.98\textwidth]{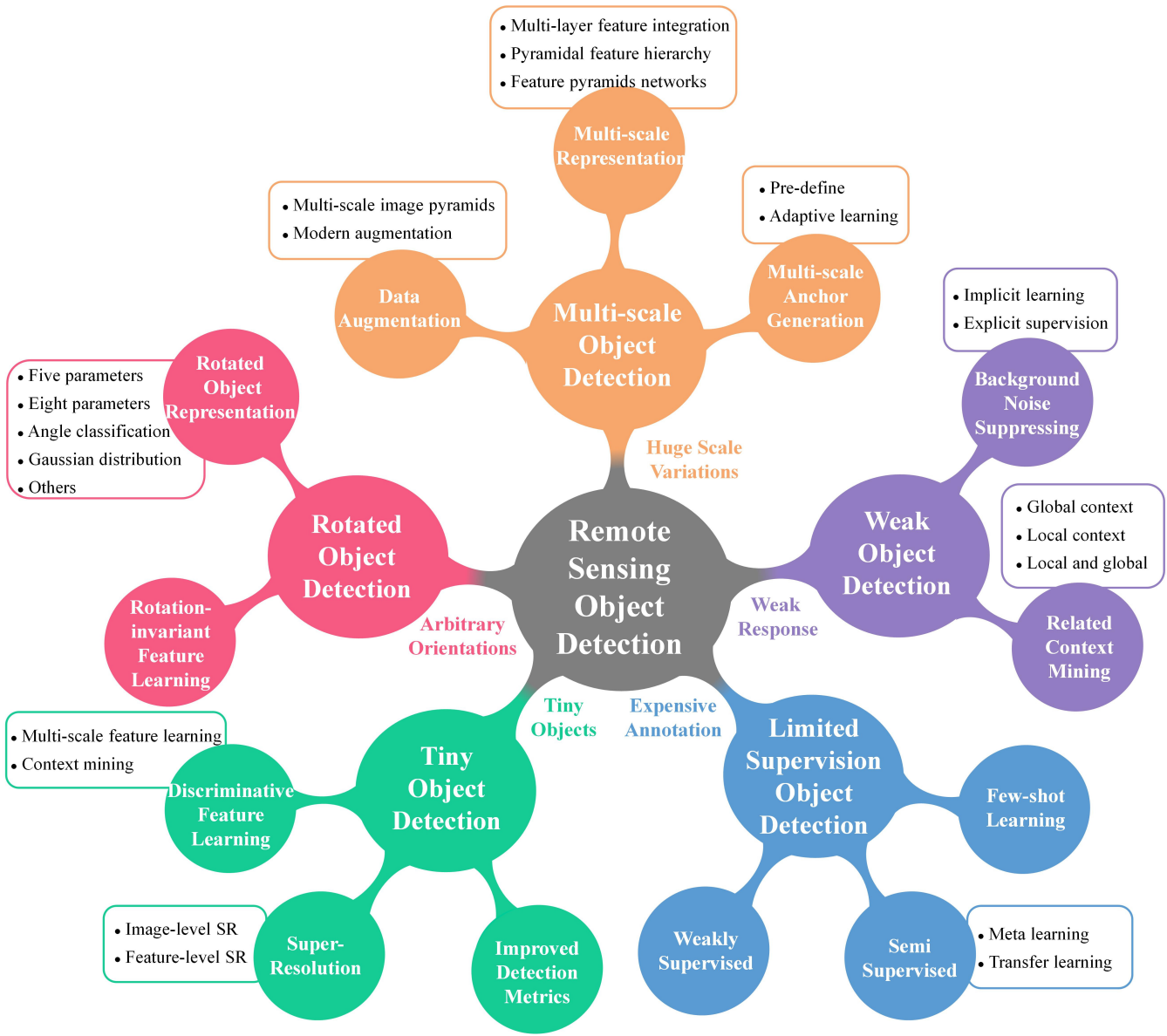}
		\end{center}
		\vspace{-3mm}
		\caption{Structured taxonomy of the deep learning based RSOD methods in this review. A hierarchical division is adopted to detailed describe each sub-category.}
		\label{fig:taxonomy}
	\end{figure*}

	Along with the prevalence of RSOD, a number of geospatial object detection surveys \cite{DIOR, cheng2016survey, weak-reponse-survey, airplane-survey, RS-Survey, Access-survey, DOTAv2, FAIR1M} have been published in recent years. For example, Cheng \textit{et al.} \cite{cheng2016survey} reviewed the early development of RSOD. Han \textit{et al.} \cite{weak-reponse-survey} focused on small and weak object detection in RSIs. In \cite{airplane-survey}, the authors reviewed airplane detection methods. Li \textit{et al.} \cite{RS-Survey} conducted a thorough survey on deep learning based detectors in the remote sensing community according to various improvement strategies. Besides, some work \cite{DIOR, DOTAv2, FAIR1M} mainly focused on publishing novel benchmark datasets for RSOD and briefly reviewed object detection methods in the field of remote sensing. Compared with previous works, this survey provides a comprehensive analysis of the major challenges in RSOD based on the characteristics of geospatial objects and systematically categorizes and summarizes the deep learning based remote sensing object detectors according to these challenges. Moreover, more than 300 papers on RSOD are reviewed in this work, leading to a more comprehensive and systematic survey.
	
	Fig. \ref{fig:taxonomy} shows the taxonomy of object detection methods in this review. According to the major challenges in RSOD, we divide the current deep learning based RSOD methods into five main categories: multi-scale object detection, rotated object detection, weak object detection, tiny object detection, and object detection with limited supervision. In each category, we further summarize the sub-categories based on the improvement strategies or learning paradigms designed for the category-specific challenges. For multi-scale object detection, we mainly review the three widely used methods: data augmentation strategy, multi-scale feature representation, and high-quality multi-scale anchor generation. With regard to rotated object detection, we mainly focus on the rotated bounding box representation and rotation-insensitive feature learning. For weak object detection, we divide it into two classes: background noise suppressing and related context mining. As for tiny object detection, we detail it into three streams: discriminative feature extraction, super-resolution reconstruction, and improved detection metrics. According to the learning paradigms, we divide object detection with limited supervision into weakly-supervised object detection, semi-supervised object detection, and few-shot object detection. Notably, there are still detailed divisions in each sub-category, as shown in the rounded rectangles in Fig. \ref{fig:taxonomy}. This hierarchical division provides a systematic review and summarization of existing methods. It helps researchers understand RSOD more comprehensively and facilitate further progress, which is the main purpose of this review.
	
	In summary, the main contributions of this review are as follows:
	\begin{itemize}
		\item We comprehensively analyze the major challenges in RSOD based on the characteristics of geospatial objects, including huge scale variations, arbitrary orientations, weak feature responses, tiny objects, and expensive annotations.
		
		\item We systematically summarize the deep learning based object detectors in the remote sensing community and categorize them in a hierarchical manner according to their motivation.
		
		\item We present a forward-looking discussion of future research directions for RSOD to motivate the further progress of RSOD.		
	\end{itemize}
	
	\section{Multi-scale Object Detection}
	\begin{figure}[t!]
		\begin{center}
			\includegraphics[width=0.48\textwidth]{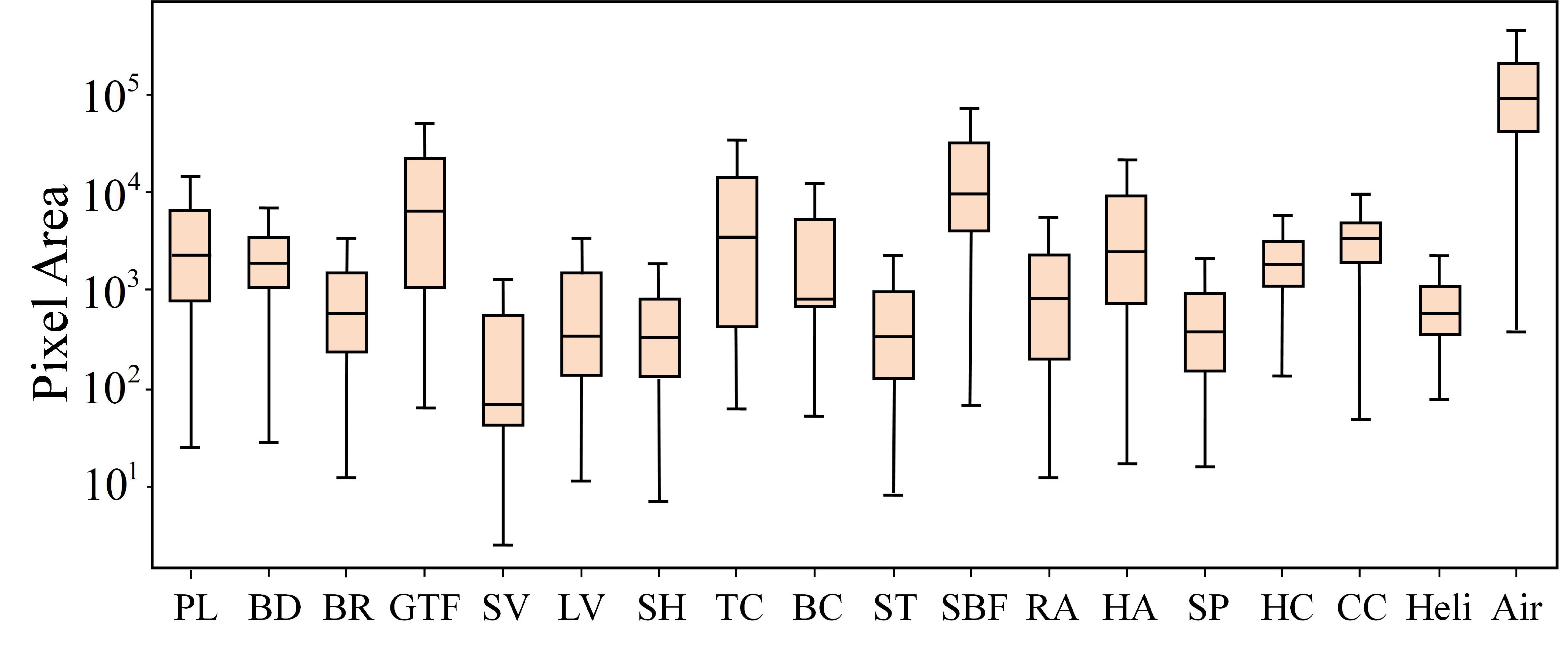}
		\end{center}
		\vspace{-3mm}
		\caption{Scale variations for each category (the short names for categories can be referred to \cite{DOTAv2}) in the DOTAv2.0 dataset. Huge scale variations exist in both inter-categories and intra-categories.}
		\label{fig:DOTA-Boxplot}
\end{figure}
\begin{figure}[t!]
		\begin{center}
			\includegraphics[width=0.47\textwidth]{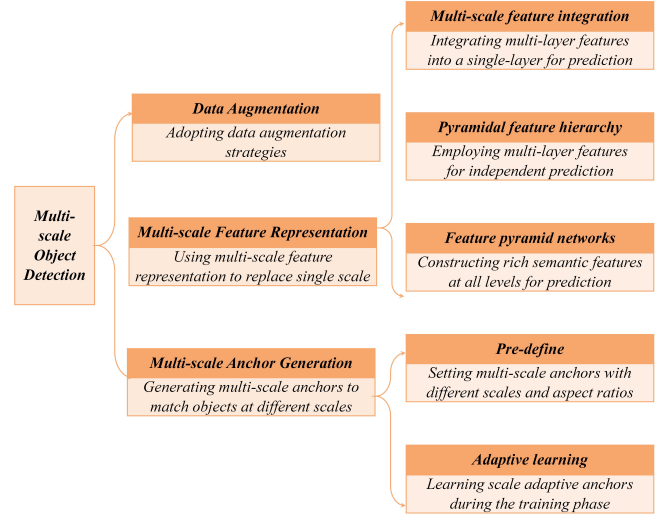}
		\end{center}
		\vspace{-3mm}
		\caption{A brief summary of multi-scale object detection methods.}
		\label{fig:MSOD}
	\end{figure}

	Due to the different spatial resolutions between RSIs, the huge scale variation is a notoriously challenging problem in RSOD and seriously degrades the detection performance. As depicted in Fig. \ref{fig:DOTA-Boxplot}, we present the distribution of object pixel areas for each category in the DOTAv2.0 dataset \cite{DOTAv2}. Obviously, the scales vary greatly between categories, in which a small vehicle may only contain less than 10 pixel area while an airport exceeds 10$^{5}$ pixel area. Worse still, the huge intra-category scale variations further exacerbate the difficulties of multi-scale object detection. To tackle the huge scale variation problem, current studies are mainly divided into data augmentation, multi-scale feature representation, and multi-scale anchor generation.	
	Fig. \ref{fig:MSOD} gives a brief summary of multi-scale object detection methods.

	\subsection{Data Augmentation}
	Data augmentation is a simple yet widely applied approach for increasing dataset diversity. As for the scale variation problem in multi-scale object detection, image scaling is a straightforward and effective augmentation method. Zhao \textit{et al.} \cite{Image-block} fed multi-scale image pyramids into multiple networks and fused the output features from these networks to generate multi-scale feature representations. In \cite{ICN}, Azimi \textit{et al.} proposed a combined image cascade and feature pyramid network to extract object features on various scales. Although image pyramids can effectively increase the detection performance for multi-scale objects, the inference time and computational complexity are severely increased. To tackle this problem, Shamsolmoali \textit{et al.} \cite{LIPM} designed a lightweight image pyramid module (LIPM). The proposed LIPM receives multiple down-sampling images to generate multi-scale feature maps and fuses the output multi-scale feature maps with the corresponding scale feature maps from the backbone. Moreover, some modern data augmentation methods (e.g., Moscia and Stitcher \cite{Stitcher}) also show remarkable effectiveness in multi-scale object detection, especially for small objects \cite{Aug1,Aug2,Aug3}.
	\begin{figure*}[t!]
		\begin{center}
			\includegraphics[width=0.98\textwidth]{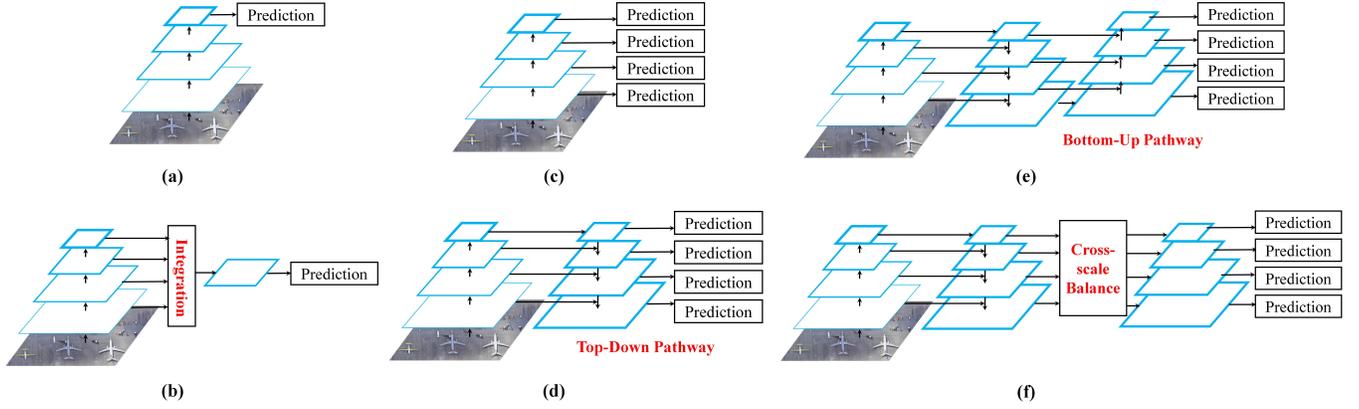}
		\end{center}
		\vspace{-3mm}
		\caption{Single-scale feature representation and six paradigms for multi-scale feature representation. (a) Single-scale feature representation. (b) Multi-scale feature integration. (c) Pyramidal feature hierarchy. (d) Feature pyramid networks. (e) Top-down and Bottom-up.  (f) Cross-scale feature balance.}
		\label{fig:Multi-scale-Feature-Representation}
	\end{figure*}
	\subsection{Multi-scale Feature Representation}
	Early studies in RSOD usually utilize the last single feature map of the backbone to detect objects, as illustrated in Fig.\ref{fig:Multi-scale-Feature-Representation}(a). However,  such a single-scale feature map prediction limits the detector to handle the object with a wide range of scale \cite{han2017efficient,RSOD,zhong2018multi}. Consequently, multi-scale feature representation methods have been proposed and have become an effective solution to the huge object scale variation problem in RSOD. The current multi-scale feature representation methods are mainly divided into three streams: multi-scale feature integration, pyramidal feature hierarchy, and feature pyramid networks.
	\subsubsection{Multi-scale Feature Integration}
	Convolutional neural networks (CNN) usually adopt a deep hierarchical structure where different level features have different characteristics. The shallow-level features usually contain fine-grained features (e.g., points, edges, and textures of objects) and provide detailed spatial location information, which is more suitable for object localization. In contrast, features from higher-level layers show stronger semantic information and present discriminative information for object classification. To combine the information from different layers and generate the multi-scale representation, some researchers introduced multi-layer feature integration methods that integrate features from multiple layers into a single feature map and perform the detection on this rebuilt feature map \cite{ISPRS-Ding,TGRS-Liu,GRSL-Liu,Hierarchical-TGRS,GRSL-L2-Norm,ISPRS-Deng,HyNet,deformable-Faster-RCNN-RS}. Fig. \ref{fig:Multi-scale-Feature-Representation}(b) depicts the structure of multi-layer feature integration methods.
		
	Zhang \textit{et al.} \cite{Hierarchical-TGRS} designed a hierarchical robust CNN to extract hierarchical spatial semantic information by fusing multi-scale convolutional features from three different layers and introduced multiple fully connected layers to enhance the rotation and scaling robustness of the network. Considering the different norms between multi-layer features, Lin \textit{et al.} \cite{GRSL-L2-Norm} applied an L2 normalization for each feature before integration to maintain stability in the network training stage. Unlike previous multi-scale feature integration at the level of the convolutional layer, Zheng \textit{et al.} \cite{HyNet} designed the HyBlock to build multi-scale feature representation at the intra-layer level. The HyBlock employs the atrous separable convolution with pyramidal receptive fields to learn the hyper-scale features, alleviating the scale-variation issue in RSOD.
	\subsubsection{Pyramidal Feature Hierarchy}
	The key insight behind the pyramidal feature hierarchy is that the features in different layers can encode object information from different scales. For instance, small objects are more likely to appear in shallow layers, while large objects tend to exist in deep layers. Therefore, the pyramidal feature hierarchy employs multiple-layer features for independent prediction to detect objects with a wide scale range, as depicted in Fig. \ref{fig:Multi-scale-Feature-Representation}(c). SSD \cite{SSD} is a typical representative of the pyramidal feature hierarchy, which has a wide range of extended applications in both natural scenes \cite{RFBNet,DSOD,SSDES} and remote sensing scenes \cite{Attention-Feature-Fusion-SSD, Full-Scale, TGRS-Houbiao,TCSFT,SSD-Dulan,SSD-Access,SSD-TGRS}.
	
	To improve the detection performance for small vehicles, Liang \cite{TCSFT} \textit{et al.} added an extra scaling branch to the SSD, which consists of a deconvolution module and an average pooling layer. Referring to hierarchical regression layers in SSD, Wang \textit{et al.} \cite{Full-Scale} introduced the scale-invariant regression layers (SIRLs), where three isolated regression layers are employed to capture the information of full-scale objects. Based on the SIRLs, a novel specific scale joint loss is introduced to accelerate network convergence. In \cite{TGRS-Moulichao}, Li \textit{et al.} proposed the HSF-Net that introduces the hierarchical selective filtering layer in both RPN and detection sub-network. Specifically, the hierarchical selective filtering layer employs three convolutional layers with different kernel sizes (e.g., $1\times1$, $3\times3$, and $5\times5$) to obtain multiple receptive field features, which benefits multi-scale ship detection.
		
	\subsubsection{Feature Pyramid Networks}
	Pyramidal feature hierarchy methods use independent multi-level features for detection and ignore the complementary information between features at different levels, resulting in weak semantic information for low-level features. To tackle this problem, Lin \textit{et al.} \cite{FPN} proposed the feature pyramid network (FPN). As shown in Fig. \ref{fig:Multi-scale-Feature-Representation}(d), the FPN introduces a top-down pathway to transfer the rich semantic information from high-level features to shallow-level features, leading to rich semantic features at all levels (please refer to the detailed in \cite{FPN}). Thanks to the significant improvement of FPN for multi-scale object detection, FPN and its extensions \cite{PANet,LibraR-CNN,efficientdet} play a dominant role in multi-scale feature representation.
	
	Considering the extreme aspect ratios of geospatial objects (e.g., bridges, harbors, and airports), Hou \textit{et al.} \cite{TIP-FPN} proposed an asymmetric feature pyramid network (AFPN). The AFPN adopts the asymmetric convolution block to enhance the feature representation regarding the cross-shaped skeleton and improve the performance of large aspect ratio objects. Zhang \textit{et al.} \cite{Laplacian-FPN} designed a Laplacian feature pyramid network (LFPN) to inject high-frequency information into the multi-scale pyramidal feature representation, which is useful for accurate object detection but has been ignored by previous work. In \cite{HR-FPN-RS}, Zhang \textit{et al.} introduced the high-resolution feature pyramid network (HRFPN) to fully leverage the high-resolution feature representations, leading to precise and robust SAR ship detection. In addition, some researchers integrated the novel feature fusion module \cite{FPN3,FPN8}, attention machine \cite{FPN1,FPN2,FPN4,FPN7}, or dilation convolution layer \cite{FPN5,FPN6} into the FPN to further obtain a more discriminative multi-scale feature representation.
	
	The FPN introduces a top-down pathway to transfer the high-level semantic information into the shallow layers, while the low-level spatial information is still lost in the top layers after the long-distance propagation in the backbone. Drawing on this problem, Fu \textit{et al.} \cite{ISPRS-czh} proposed a feature-fusion architecture (FFA) that integrates an auxiliary bottom-up pathway into the FPN structure to transfer the low-level spatial information to the top layers features via a short path, as depicted in Fig. \ref{fig:Multi-scale-Feature-Representation}(e). The FFA ensures the detector extracts multi-scale feature pyramids with rich semantic and detailed spatial information. Similarly, in \cite{Bi-FPN-TGRS, Bi-FPN-PR}, the authors introduced a bidirectional FPN that learns the importance of different level features through learnable parameters and fuses the multi-level features through iteratively top-down and bottom-up pathways.
	
	Different from the above sequential enhancement pathway \cite{ISPRS-czh}, some studies \cite{GRSL-Cross-scale, Cross-scale-Fukun, Cross-scale-Attention,Cross-scale-Libra-RCNN,Cross-scale1, FPT-RS,JSTAR-Cross-scale-self-attention,Cross-scale-Nuercomputing,ISPRS-Part-based, ISPRS-Balance-Learning,FPN9, Fine-grained-detection} adopt a cross-level feature fusion manner. As shown in Fig. \ref{fig:Multi-scale-Feature-Representation}(f), the cross-level feature fusion methods fully collect the features at all levels to adaptively obtain balanced feature maps. Cheng \textit{et al.} \cite{GRSL-Cross-scale} utilized feature concatenation operation to achieve cross-scale feature fusion. Considering that features from different levels should have different contributions to the feature fusion, Fu \textit{et al.} \cite{Cross-scale-Fukun} proposed level-based attention to learn the unique contribution of features from each level. Thanks to the powerful global information extraction ability of the transformer structure, some work \cite{FPT-RS, JSTAR-Cross-scale-self-attention} introduced the transformer structures to integrate and refine multi-level features. In \cite{Cross-scale-Nuercomputing}, Chen \textit{et al.} presented a cascading attention network where position supervision is introduced to enhance the semantic information of multi-level features.
	\subsection{Multi-scale Anchor Generation}
	Apart from the data augmentation and multi-scale feature representation methods, multi-scale anchor generation can also tackle the huge object scale variation problem in RSOD. Due to the difference in the scale range of objects in natural and remote sensing scenes, some studies \cite{Anchor-RS-Guo,Anchor-GRSL-Zhang,Anchor-ICIP-Yang,Class-specific-anchor-TGRS-Dong,Aspect-Ratio-Attention, Self-Adaptive-Aspect-Ratio, Class-specific-Anchor-RS, Guide-anchor-RS, Guide-anchor-TGRS, Guide-anchor-ISPRS} modify the anchor settings in common object detection to better cover the scales of geospatial objects. 
	\begin{figure}[t!]
		\begin{center}
			\includegraphics[width=0.47\textwidth]{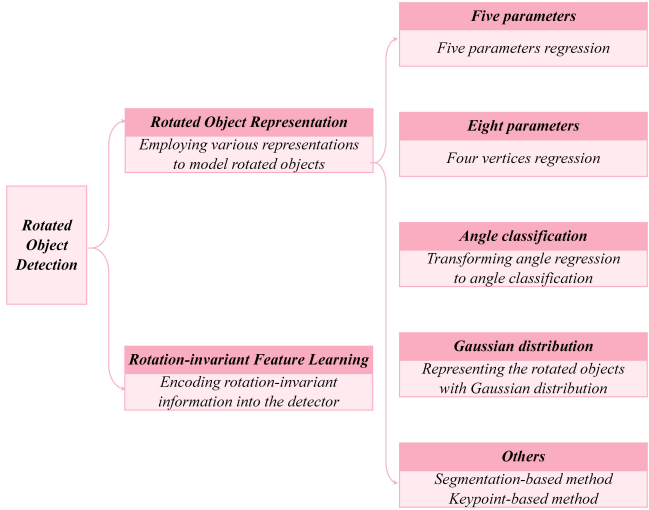}
		\end{center}
		\vspace{-3mm}
		\caption{A brief summary of rotated object detection methods.}
		\label{fig:Rotated-OD}
	\end{figure}

	Guo \textit{et al.} \cite{Anchor-RS-Guo} injected extra anchors with more scales and aspect ratios into the detector for multi-scale object detection. Dong \textit{et al.} \cite{Class-specific-anchor-TGRS-Dong} designed more suitable anchor scales based on the statistics of object scales in the training set. Qiu \textit{et al.} \cite{Aspect-Ratio-Attention} extended the original square RoI features into vertical, square, and horizontal RoI features and fused these RoI features to represent objects with different aspect ratios in a more flexible way. The above methods follow a fixed anchor setting, while current studies \cite{Self-Adaptive-Aspect-Ratio, Class-specific-Anchor-RS, Guide-anchor-RS, Guide-anchor-TGRS, Guide-anchor-ISPRS} have attempted to dynamically learn the anchor during the training phase. Considering the aspect ratio variations between different categories, Hou \textit{et al.} \cite{Self-Adaptive-Aspect-Ratio} devised a novel self-adaptive aspect ratio anchor (SARA) to adaptively learn an appropriate aspect ratio for each category. The SARA embeds the learnable category-wise aspect ratio values into the regression branch to adaptively update the aspect ratio of each category with the gradient of the location regression loss. Inspired by GA-RPN \cite{GA-RPN}, some researchers \cite{Guide-anchor-RS, Guide-anchor-TGRS, Guide-anchor-ISPRS} introduced a lightweight sub-network into the detector to adaptively learn the location and shape information of anchors.
	\section{Rotated Object Detection}
	Arbitrary orientation of objects is another major challenge in RSOD. Since the objects in RSIs are acquired from a bird's eye view, they exhibit the property of arbitrary orientations, so the widely used horizontal bounding box (HBB) representation in generic object detection is insufficient to locate rotated objects accurately. Therefore, numerous researchers have focused on the arbitrary orientation property of geospatial objects, which can be summarized into rotated object representation and rotation-invariant feature learning.
	A brief summary of rotated object detection methods is depicted in Fig. \ref{fig:Rotated-OD}.
	\begin{figure}[t!]
	\centering
	\includegraphics[width=0.48\textwidth]{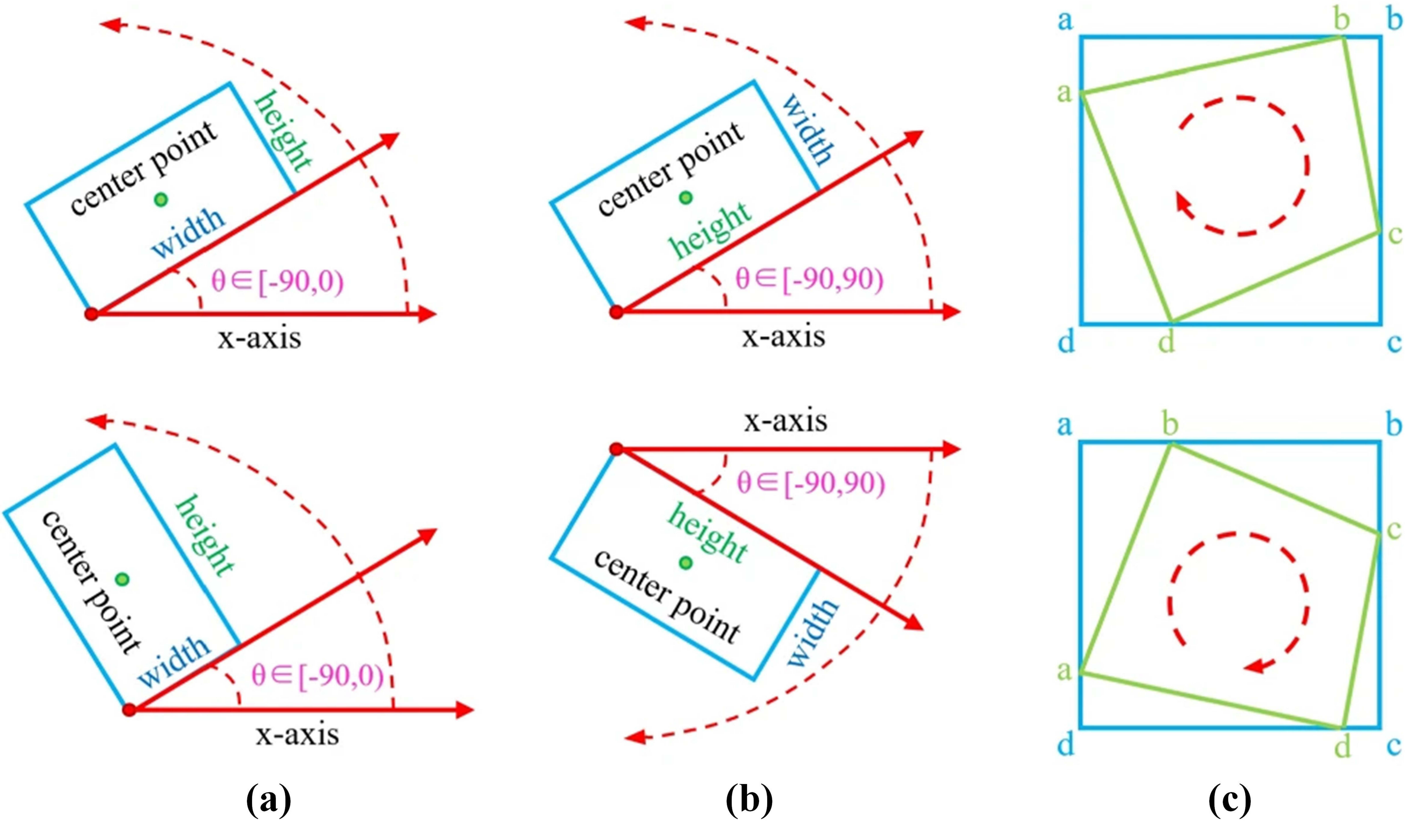}
	\vspace{-3mm}
	\caption{Visualization of the five parameters representation and eight parameters representation methods for rotated objects \cite{RO13}.}
	\label{fig. RO-parameters}
	\end{figure}
	\subsection{Rotated Object Representation}
	Rotated object representation is essential for RSOD to avoid redundant backgrounds and obtain precise detection results. Recent rotated object representation methods can be mainly summarized into several categories: five parameters representation \cite{RO1,RO2,RO3,Access-Yangxue, CFC-Net, DAL, RO4,RO5, DRBoxv2, R3Net}, eight parameters representation \cite{DOTAv1,RO7,RO8,RO9,RO10,RO11,RO12,Oriented-RCNN,Oriented-RCNN-v2,RIL-Mingqi}, angle classification representation \cite{RO15,RO16,RO13,RO14}, gaussian distribution representation \cite{RO19,RO20,RO17,RO18}, and others \cite{RO21,RO23,Ship-Tangxu,RO22, RO24, RO25, RO26, RO28,RO30,Polar-Coordinates,AProNet}.
	
	\subsubsection{Five Parameters}
	The most popular solution is representing objects with a five-parameter method $(x, y, w, h, \theta)$, which simply adds an extra rotation angle parameter $\theta$ on HBB \cite{RO1,RO2,RO3,Access-Yangxue, CFC-Net, DAL, RO4,RO5, DRBoxv2}. The definition of the angular range plays a crucial role in such methods, where two kinds of definitions are derived. Some studies \cite{RO1,RO2,RO3,Access-Yangxue,CFC-Net, DAL} define $\theta$ as the acute angle to the x-axis and restrict the angular range to 90$^{\circ}$, as shown in Fig. \ref{fig. RO-parameters}(a). As the most representative work, Yang \textit{et al.} \cite{RO1} followed the five parameters method to detect rotated objects and designed an IoU-aware loss function to tackle the boundary discontinuity problem of rotation angles. Another group \cite{RO4,RO5,DRBoxv2,R3Net} refers to $\theta$ as the angle between the x-axis and the long side, whose range is 180$^{\circ}$, as depicted in Fig. \ref{fig. RO-parameters}(b). Ding \textit{et al.} \cite{RO5} regressed rotation angles by five-parameter methods and transformed the features of horizontal regions into rotated ones to facilitate rotated object detection. 
	\begin{figure}[t!]
		\centering
		\subfloat{
			\begin{minipage}[b]{.23\textwidth}
				\centering
				\includegraphics[width=0.9\linewidth]{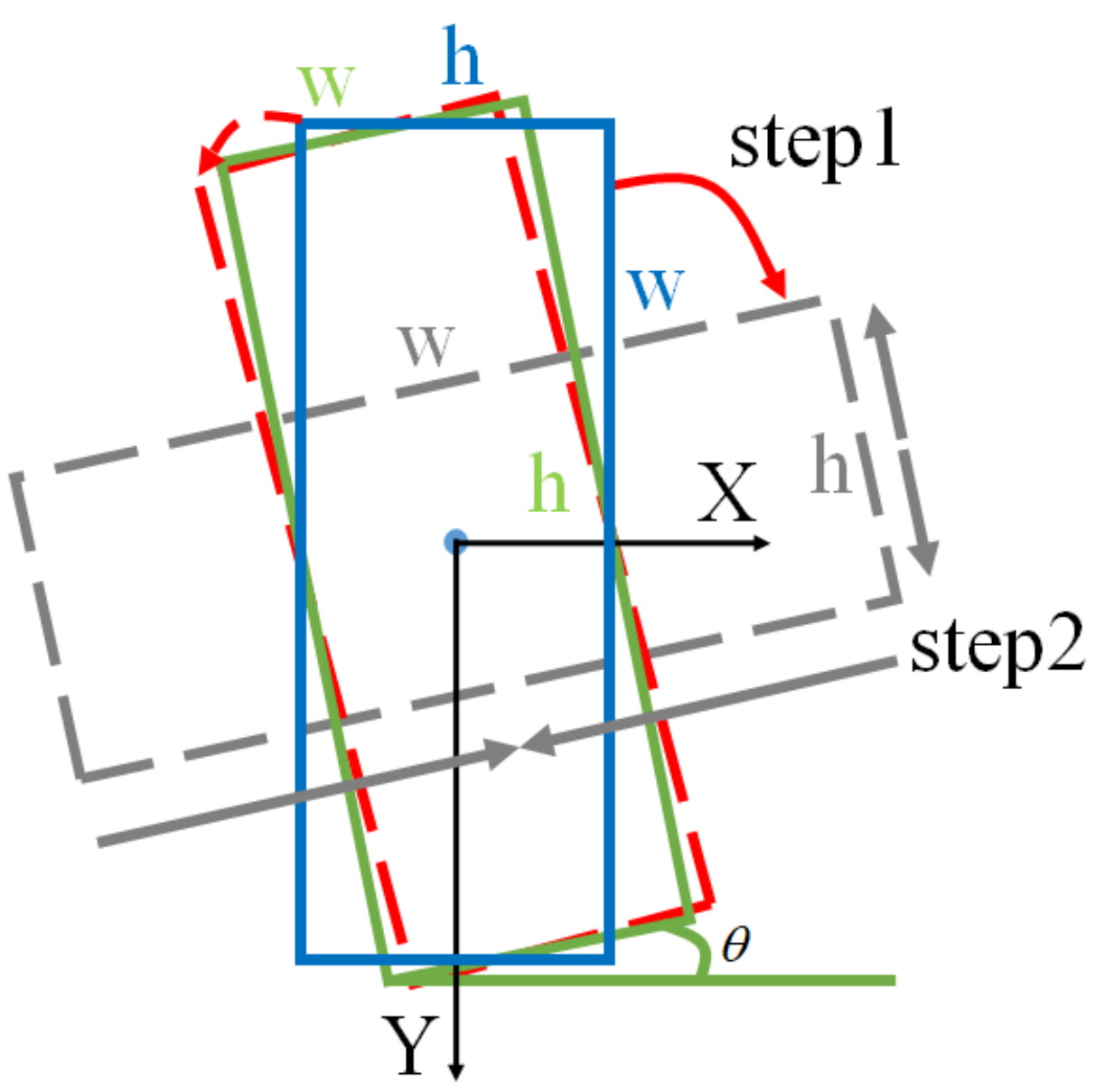}		
			\end{minipage}%
		}%
		\subfloat{
			\begin{minipage}[b]{.23\textwidth}
				\centering
				\includegraphics[width=0.9\linewidth]{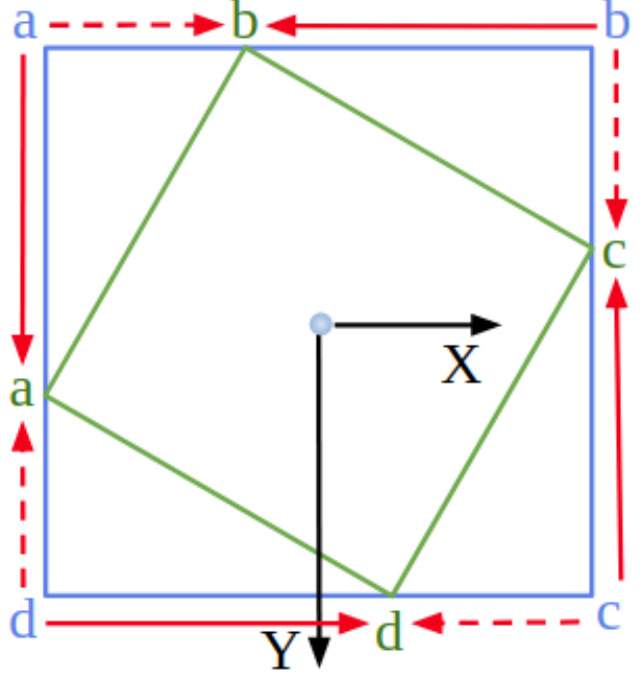}		
			\end{minipage}%
		}%
		\vspace{-1mm}
		\caption{Boundary discontinuity challenge of five parameters method and eight parameters method \cite{RO8,RO10}. }
		\label{fig. Boundary}
	\end{figure}	
	\subsubsection{Eight Parameters} Different from five-parameter methods, eight-parameter methods \cite{DOTAv1,RO7,RO8,RO9,RO10,RO11,RO12,Oriented-RCNN,Oriented-RCNN-v2,RIL-Mingqi} solve the issue of rotated object representation by directly regressing four vertices $\{(a_x,a_y),(b_x,b_y),(c_x,c_y),(d_x,d_y)\}$, as shown in Fig. \ref{fig. RO-parameters}(c). Xia \textit{et al.} \cite{DOTAv1} first adopted the eight-parameter method for rotated object representation, which directly supervises the detection model by minimizing the difference between each vertex and ground truth coordinates during training. However, the sequence order of these vertices is essential for the eight-parameter method to avoid unstable training. As shown in Fig. \ref{fig. Boundary}, it is intuitive that regressing targets from the red dotted arrow is an easier route, but the actual process follows the red solid arrows, which causes the difficulty of model training. To this end, Qian \textit{et al.} \cite{RO8,RO10} proposed a modulated loss function that calculates the losses under different sorted orders and selects the minimum case to learn, efficiently improving the detection performance.
	\subsubsection{Angle Classification}
	To address the issue described in Fig. \ref{fig. Boundary} from the source, many researchers \cite{RO13,RO14,RO15,RO16} take a detour from the boundary challenge of regression by transforming the angle prediction problem into an angle classification task. Yang \textit{et al.} \cite{RO13} proposed the first angle classification method for rotated object detection, which converts the continuous angle into a discrete kind and trains the model with novel circular smooth labels. However, the angle classification head \cite{RO13} introduces additional parameters and degrades the detector's efficiency. To overcome this, Yang \textit{et al.} \cite{RO14} improved the \cite{RO13} with a densely coded label that ensures both the accuracy and efficiency of the model.

	\subsubsection{Gaussian Distribution}
\begin{figure}[t]
		\centering
		\includegraphics[width=0.46\textwidth]{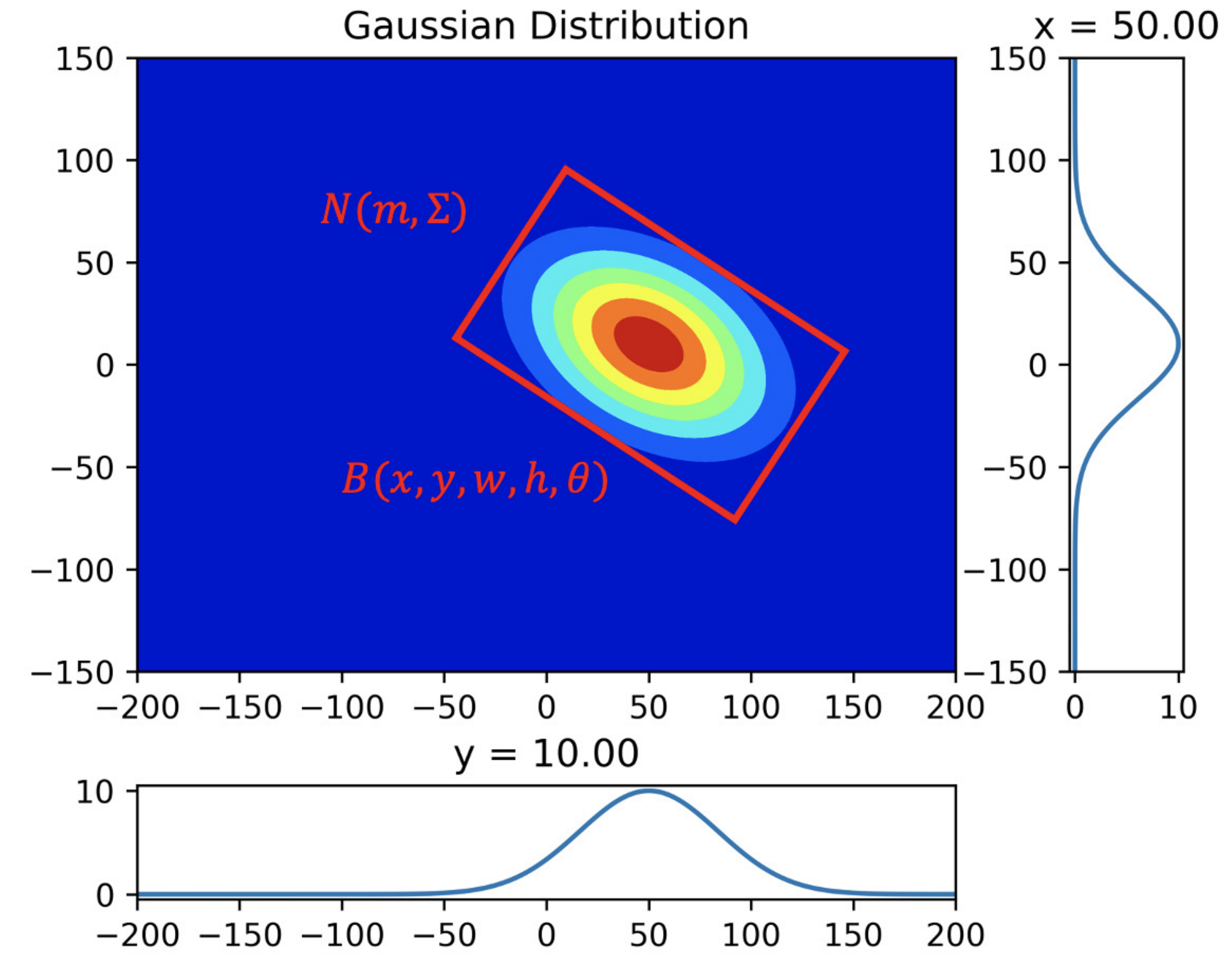}
		\vspace{-1mm}
		\caption{Visualization of the Gaussian distribution representation methods for rotated objects \cite{RO19}.}
		\label{fig. RO-Gaussian}
	\end{figure}

	Although the above methods achieve promising progress, they do not consider the misalignment between the actual detection performance and optimization metric. Most recently, a series of works \cite{RO19,RO20,RO17,RO18} aims to handle this challenge by representing rotated objects with Gaussian distribution, as shown in Fig. \ref{fig. RO-Gaussian}. Specifically, these methods convert rotated objects into a 2D Gaussian distribution $\mathcal{N}\left({\mu}, {\Sigma}\right)$ as follows:
	\begin{equation}
	\begin{aligned}
	{\Sigma}^{1 / 2} &=\mathbf{R} \mathbf{\Lambda} \mathbf{R}^{\top} \\
	&=\left(\begin{array}{cc}
	\cos \theta & -\sin \theta \\
	\sin \theta & \cos \theta
	\end{array}\right)\left(\begin{array}{cc}
	\frac{w}{2} & 0 \\
	0 & \frac{h}{2}
	\end{array}\right)\left(\begin{array}{cc}
	\cos \theta & \sin \theta \\
	-\sin \theta & \cos \theta
	\end{array}\right) \\
	&=\left(\begin{array}{cc}
	\frac{w}{2} \cos ^{2} \theta+\frac{h}{2} \sin ^{2} \theta & \frac{w-h}{2} \cos \theta \sin \theta \\
	\frac{w-h}{2} \cos \theta \sin \theta & \frac{w}{2} \sin ^{2} \theta+\frac{h}{2} \cos ^{2} \theta
	\end{array}\right) \\
	{\mu} &=(x, y)^{\top}
	\end{aligned}
	\label{Eq.Gaussian-distribution}
	\end{equation}
	where $\mathbf{R}$ represents the rotation matrix, and $\mathbf{\Lambda}$ represents the diagonal matrix of eigenvalues.
	With the Gaussian distribution representation in Eq. \ref{Eq.Gaussian-distribution}, the IoU between two rotated objects can be simplified as a distance estimation between two distributions. Besides, the Gaussian distribution representation discards the definition of angular boundary and effectively solves the angular boundary problem. Yang \textit{et al.} \cite{RO19} proposed a novel metric with Gaussian Wasserstein distance (GWD) for measuring the distance between distributions, which achieves remarkable performance by efficiently approximating the rotation IoU. Based on this, Yang \textit{et al.} \cite{RO20} introduced a Kullback-Leibler divergence (KLD) metric to enhance its scale invariance.
	\subsubsection{Others}
	Some researchers solve the rotated object representation by other approaches, such as segmentation-based \cite{RO21,RO23,Ship-Tangxu} and keypoint-based \cite{RO22, RO24, RO25, RO26, RO28,RO30,Polar-Coordinates,AProNet}. The representative one in segmentation-based methods is Mask OBB \cite{RO21}, which deploys the segmentation method on each horizontal proposal to obtain the pixel-level object region and produce the minimum external rectangle as a rotated bounding box. On the other side, Wei \textit{et al.} \cite{RO30} adopted a keypoint-based representation for rotated objects, which locates the object center and leverages a pair of middle lines to represent the whole object.
	In addition, Yang \textit{et al.} \cite{H2RBox} proposed the first rotated object detector supervised by horizontal box annotations, which adopts the self-supervised learning of two different views to predict the angles of rotated objects.

	\subsection{Rotation-invariant Feature Learning}	
	Rotation-invariant features indicate the features remain consistent under any rotation transformations. Thus, rotation-invariant feature learning of objects is a crucial research field to tackle the arbitrary orientation problem in rotated object detection.  
	To this purpose, many researchers proposed a series of methods for learning the rotational invariance of objects \cite{RI1,RI2,RI3,RI4,RI5,RI6,RI7,RI8,RI9,RI10,S2ANet,RI12}, which significantly improves rotated object detection in RSIs. 
	
	Cheng \textit{et al.} \cite{RI1} proposed the first rotation-invariant object detector to precisely recognize objects by using rotation-insensitive features, which enforces the features of objects to be consistent at different rotation angles. Later, Cheng \textit{et al.} \cite{RI3,RI4} employed the rotation-invariant and fisher discrimination regularizers to encourage the detector to learn both rotation-invariant and discriminative features. In \cite{RI5, RI6},  Wu \textit{et al.} analyzed object rotation invariance under polar coordinates in the Fourier domain and designed a spatial-frequency channel feature extraction module to obtain the rotation-invariant features. Considering misalignment between axis-aligned convolutional features and rotated objects, Han \textit{et al.} \cite{S2ANet} proposed an oriented detection module that adopts a novel alignment convolution operation to learn the orientation information. In \cite{RI10}, Han \textit{et al.} further devised a rotation-equivariant detector to explicitly encode rotation equivariance and rotation invariance. Besides, some researchers \cite{RI12,ISPRS-czh} extended the RPN with a series of predefined rotated anchors to cope with the arbitrary orientation characteristics of geospatial objects.
	
\begin{table}[t!]
	\renewcommand\arraystretch{1.4}
	\setlength\tabcolsep{4pt} 
	\scriptsize
	\centering		
	\caption{Detection performance of rotated object detection methods on the DOTAv1.0 dataset with rotated annotations.}
	\vspace{-3mm}
	\begin{tabular}{cccc}
		\hline
		Models &  Backbone & Methods & mAP(\%)   \\                
		\hline
		SCRDet\cite{RO1} & R-101-FPN & Five parameters & 72.61\\
		O$^{2}$Det\cite{RO30} & H-104 & Keypoint-based &72.8 \\
		R$^{3}$Det\cite{RO2} & R-101-FPN & Five parameters & 73.79\\
		S$^{2}$ANet\cite{S2ANet} & R-50-FPN & Rotation-invariant feature &74.12 \\
		RoI Transformer \cite{RO5} & R-50-FPN  & Five parameters & 74.61\\
		Mask OBB\cite{RO21} & R-50-FPN & Segmentation-based & 74.86\\
		Gliding Vertex\cite{RO9} &R-101-FPN & Four vertices & 75.02 \\
		DCL\cite{RO16} & R-152-FPN & Angle classification & 75.54 \\
		ReDet\cite{RI10} & ReR50-ReFPN & Rotation-invariant feature & 76.25 \\
		Oriented R-CNN\cite{Oriented-RCNN} & R-101-FPN & Four vertices & 76.28\\
		R$^{3}$Det-KLD\cite{RO20} & R-50-FPN & Gaussian distribution & 77.36\\
		\hline
	\end{tabular}
	\vspace{-3mm}
	\label{summary-DOTA-OBB-results}
\end{table}	
	We summarize the detection performance of milestone rotated object detection methods in Table \ref{summary-DOTA-OBB-results}.
	\section{Weak Object Detection}
\begin{figure}[t!]
		\begin{center}
			\includegraphics[width=0.47\textwidth]{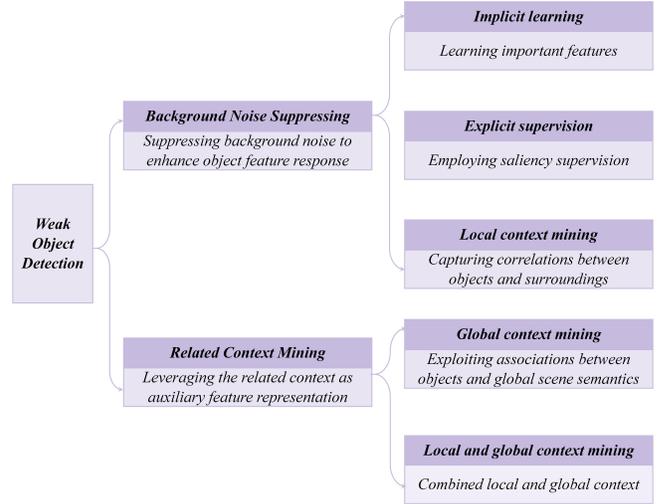}
		\end{center}
		\vspace{-3mm}
		\caption{A brief summary of weak object detection methods.}
		\label{fig:Weak-OD}
	\end{figure}

	Objects of interest in RSIs are typically embedded in complex scenes with intricate object spatial patterns and massive background noise. The complex context and background noise severely harm the feature representation of objects of interest, resulting in weak feature responses to objects of interest. Thus, many existing works have concentrated on improving the feature representation of the objects of interest, which can be grouped into two streams: suppressing background noise and mining related context information.
	A brief summary of weak object detection methods is given in Fig. \ref{fig:Weak-OD}.
	\subsection{Suppressing background noise}
	This type of method aims to strengthen the weak response of the object region in the feature map by weakening the response of background regions. It can be mainly divided into two categories: implicit learning and explicit supervision.

	\subsubsection{Implicit Learning}
	Implicit learning methods employ carefully designed modules into the detector to adaptively learn important features and suppress redundant features during the training phase, thereby reducing the background noise interference.
	
	In machine learning, dimensionality reduction can effectively learn compact feature representation and suppress irrelevant features. Drawing on the above property, Ye \textit{et al.} \cite{F3Net} proposed a feature filtration module that captures the low-dimensional feature maps by consecutive bottleneck layers to filter background noise interference. Inspired by the selective focus of human visual perception, the attention mechanism has been proposed and heavily researched\cite{SENet,SKNet,DANet}. The attention mechanism redistributes the feature importance during the network learning phase to enhance important features and suppress redundant information. Consequently, the attention mechanism has also been widely introduced in RSOD to tackle the background noise interference problem \cite{Nonlocal-Aware, SAR-Attention-TGRS-Sun,FSME,SSE-Attention,Spatial-Channel-Attention,Time-Frequency-Attention,Attention-1,Attention-2,Attention-3,Attention-Feature-Fusion-SSD}. In \cite{Nonlocal-Aware}, Huang \textit{et al.} emphasized the importance of patch-patch dependencies for RSOD and designed a novel non-local perceptual pyramidal attention (NP-Attention). The NP-Attention learns spatial multi-scale non-local dependencies and channel dependencies to enable the detector to concentrate on the object region rather than the background. Considering the strong scattering interference of the land area in SAR images, Sun \textit{et al.} \cite{SAR-Attention-TGRS-Sun} presented a ship attention module to highlight the feature representation of ships and reduce the false alarm from the land area. Moreover, a series of attention mechanisms devised for RSOD (e.g., spatial shuffle-group enhance attention \cite{SSE-Attention}, multi-scale spatial and channel-wise attention \cite{Spatial-Channel-Attention}, discrete wavelet multi-scale attention \cite{Time-Frequency-Attention}, etc.) have demonstrated their effectiveness in suppressing background noise. 

	\subsubsection{Explicit Supervision}
	Unlike implicit learning methods, the explicit supervision approach employs auxiliary saliency supervision information to explicitly guide the detector to highlight the foreground regions and weaken the background.
		
\begin{figure}[t!]
		\begin{center}
			\includegraphics[width=0.48\textwidth]{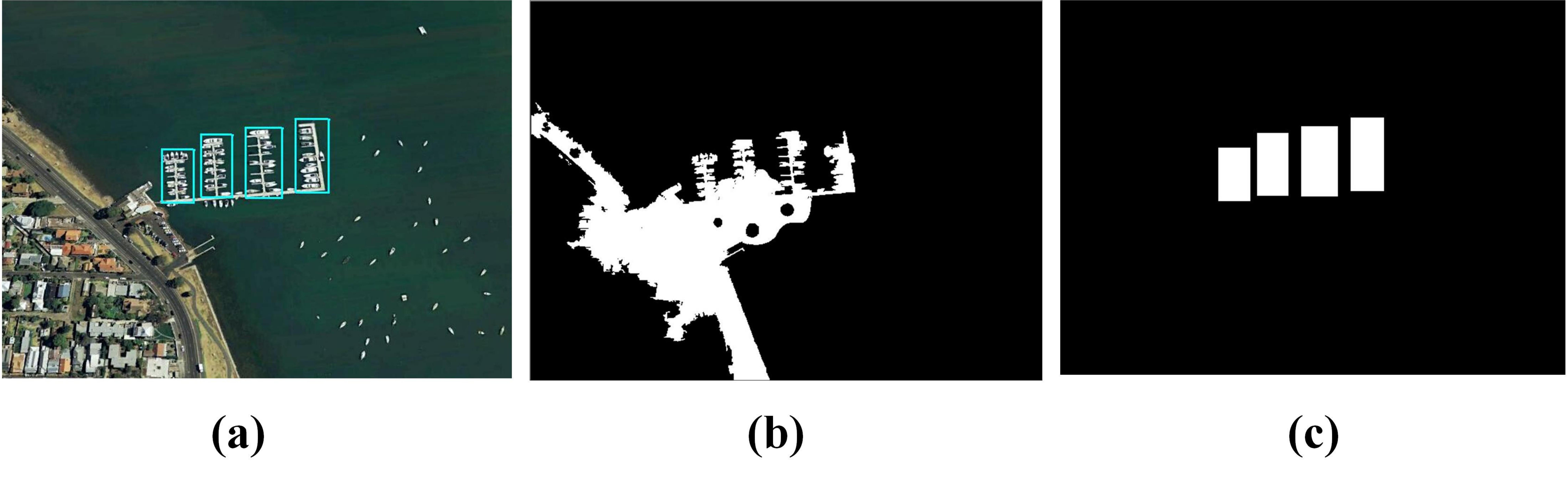}
		\end{center}
		\vspace{-5mm}
		\caption{(a) Input image. (b) Saliency map generated by the saliency detection method \cite{Random-walk-ranking}. (c) Object-level saliency map.}
		\label{fig:Supervision}
	\end{figure}

	Li \textit{et al.} \cite{Global-local-Salinecy} employed the region contrast method to obtain the saliency map and construct the saliency feature pyramid by fusing the multi-scale feature maps with the saliency map. In \cite{RECNN}, Lei \textit{et al.} extracted the saliency map with the saliency detection method \cite{Random-walk-ranking}  and proposed a saliency reconstruction network. The saliency reconstruction network utilizes the saliency map as pixel-level supervision to guide the training of the detector to strengthen saliency regions in feature maps. The above saliency detection methods are typically unsupervised, and the generated saliency map may contain non-object regions, as shown in Fig. \ref{fig:Supervision}(b), providing inaccurate guidance to the detector. Therefore, later works \cite{RO1,RO21,HSP-TGRS,FoRDet,Center-Probability,Explicit-supervison3,Explicit-supervison4,Explicit-supervison5,Explicit-supervison6} transformed the box-level annotation into object-level saliency guidance information (as shown in Fig. \ref{fig:Supervision}(c)) to generate more accurate saliency supervision. Yang \textit{et al.} \cite{RO1} designed a pixel attention network that employs object-level saliency supervision to enhance the object cues and weaken the background information. 	
	In \cite{FoRDet}, Zhang \textit{et al.} proposed the FoRDet to exploit object-level saliency supervision in a more concise way. Concretely, the proposed FoRDet leverages the prediction of foreground regions in the coarse stage (supervised under box-level annotation) to enhance the feature representation of the foreground regions in the refined stage.
	\subsection{Mining related context information }
	Context information typically refers to the spatial and semantic relations between the object and the surrounding environment or scene. This context information can provide auxiliary feature representations to the object that could not be clearly distinguished. Thus, mining context information can effectively solve the weak feature responses problem in RSOD. According to the category of context information, existing methods are mainly classified into local and global context information mining.	 
	\subsubsection{Local Context Information Mining} 
	Local context information refers to the correlation between the object and its surrounding environment in visual information and spatial distribution\cite{Local-Zhang, Context-Aware, RI2, Local-GRU, Local-graph1, Local-graph2, Local-GRSL, Local-KCPNet}. Zhang \textit{et al.} \cite{Local-Zhang} generated multiple local context regions by scaling the original region proposal into three different sizes and proposed a contextual bidirectional enhancement module to fuse the local context features with object features. The context-aware convolutional neural network (CA-CNN)\cite{Context-Aware} employed a Context-RoI mining layer to extract context information surrounding objects. The Context-RoI for an object is first generated by merging a series of filtered proposals around the object and then fused with the object RoI as the final object feature representation for classification and regression. In \cite{Local-GRU}, Ma \textit{et al.} exploit the  gated recurrent units (GRUs) to fuse object features with the local context information, leading to a more discriminative feature representation for the object. Graph Convolutional Networks (GCN) have recently shown better performance in object-object relationship reasoning. Hence, Tian \textit{et al.} \cite{Local-graph1, Local-graph2} constructed spatial and semantic graphs to model and learn the contextual relationships between objects.
	\subsubsection{Global Context Information Mining} 
	Global context information exploits the association between the object and the scene\cite{RS-Scene-Contextual,Global-HouBiao,Global-Context-Weaving,Global-Contex-Augmented,TGRS-Scene-Relevant,Scene-Vehicle,Global-1,Global-2}, e.g., vehicles generally locate on roads and ships typically appear at sea. Chen \textit{et al.} \cite{RS-Scene-Contextual} extracted the scene context information from the global image feature with the RoI-Align operation and fused it with the object-level RoI features to strengthen the relationship between the object and the scene. Liu \textit{et al.} \cite{TGRS-Scene-Relevant} designed a scene auxiliary detection head that exploits the scene context information under scene-level supervision. The scene auxiliary detection head embeds the predicted scene vector into the classification branch to fuse the object-level features with scene-level context information. In \cite{Scene-Vehicle}, Tao \textit{et al.} presented a scene context-driven vehicle detection approach. Specifically, a pre-trained scene classifier is introduced to classify each image patch into three scene categories, then the scene-specific vehicle detectors are employed to achieve preliminary detection results, and finally the detection results are further optimized with the scene contextual information.
	
	Considering the complementarity of local and global context information, Zhang \textit{et al.} \cite{CADNet} proposed a CAD-Net to mine both local and global context information. CAD-Net employed a pyramid local context network to learn the object-level local context information and designed a global context network to extract scene-level global context information. In \cite{Guide-anchor-TGRS}, Teng \textit{et al.} proposed a GLNet to collect context information from global to local so as to achieve a robust and accurate detector for RSIs. Besides, some studies \cite{DCLNet, Automated-ship-detection,FMSSD} also introduced the ASPP \cite{deeplab} or RFB module \cite{RFBNet} to leverage both local and global context information.
	\section{Tiny Object Detection }
\begin{figure}[t!]
		\begin{center}
			\includegraphics[width=0.47\textwidth]{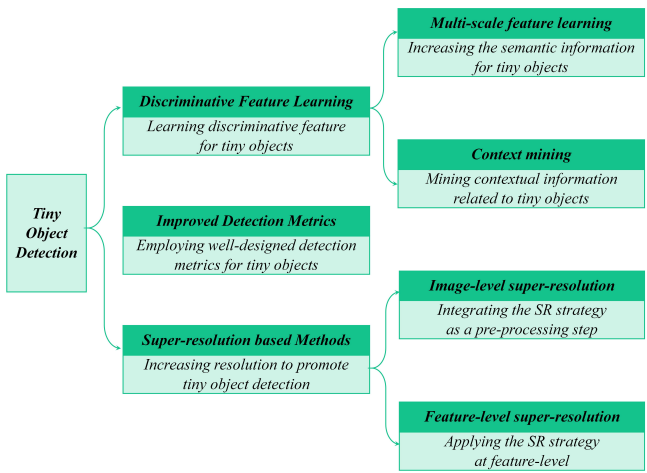}
		\end{center}
		\vspace{-3mm}
		\caption{A brief summary of tiny object detection methods.}
		\label{fig:Tiny-OD}
	\end{figure}

	The typical ground sampling distance (GSD) for RSIs is 1-3 meters, which means that even large objects (e.g., airplanes, ships, and storage tanks) can only occupy less than $16 \times 16$ pixels. Besides, even in high-resolution RSIs with a GSD of 0.25m, a vehicle with a dimension of $3 \times 1.5$m$^{2}$ only covers 72 pixels ($12 \times 6$). This prevalence of tiny objects in RSIs further increases the difficulty of RSOD. Current studies on tiny object detection are mainly grouped into discriminative feature learning, super-resolution based methods, and improved detection metrics.
	The tiny object detection methods are briefly summarized in Fig. \ref{fig:Tiny-OD}.
	\subsection{Discriminative Feature Learning}
	The extremely small scales (less than $16 \times 16$ pixels) of tiny objects make them exhibit limited appearance information, which poses serious challenges for detectors to learn the features of tiny objects. To tackle the above problem, many researchers focus on improving the discriminative feature learning ability for tiny objects\cite{Discriminative-Features1,Discriminative-Features2,Discriminative-Features3,Discriminative-Features4,Discriminative-Features5,Discriminative-Features6,Discriminative-Features7,Discriminative-Features8}.
	
	Since tiny object mainly exists in shallow features and lacks high-level semantic information \cite{FPN}, some literature \cite{Discriminative-Features1, Discriminative-Features2,Discriminative-Features3} introduces top-down structures to fuse high-level semantic information into shallow features to strengthen the semantic information for tiny objects. Considering the limited appearance information of the tiny objects, some studies \cite{Discriminative-Features4,Discriminative-Features5,Discriminative-Features6,Discriminative-Features7,Discriminative-Features8} establish the connection between the tiny object and the surrounding contextual information through the self-attention mechanism or dilated convolution to enhance the feature discriminative of tiny objects. Notably, some previously mentioned studies on multi-scale feature learning and context information mining also demonstrate remarkable effectiveness in tiny object detection. 

	\subsection{Super-resolution based Method}
	The extremely small scale is  a crucial issue for tiny object detection, so increasing the resolution of images is an intuitive solution to promote the detection performance of tiny objects. Some methods \cite{SR1,SR2,SR3,Point-to-Region} employ the super-resolution strategies as a pre-processing step into the detection pipeline to enlarge the resolution of input images. For example, Rabbi \textit{et al.} \cite{SR3} emphasized the importance of edge information for tiny object detection and proposed an edge-enhanced super-resolution generative adversarial network (GAN) to generate visually pleasing high-resolution RSIs with detailed edge information. Wu \textit{et al.} \cite{Point-to-Region} developed a Point-to-Region detection framework for tiny objects. The Point-to-Region framework first obtains the proposal regions with key-point prediction and then employs a multi-task GAN to perform the super-resolution on the proposal regions and detect the tiny objects in these proposal regions. However, the high-resolution image generated by super-resolution brings extra computational complexity to the detection pipeline. Drawing on this problem, \cite{Feature-Level-SR1} and \cite{Feature-Level-SR2} employ the super-resolution strategy at the feature level to acquire discriminative feature representation for tiny objects and effectively save computational resources.
\begin{figure}[t!]
	\begin{center}
		\centering
		\includegraphics[width=0.48\textwidth]{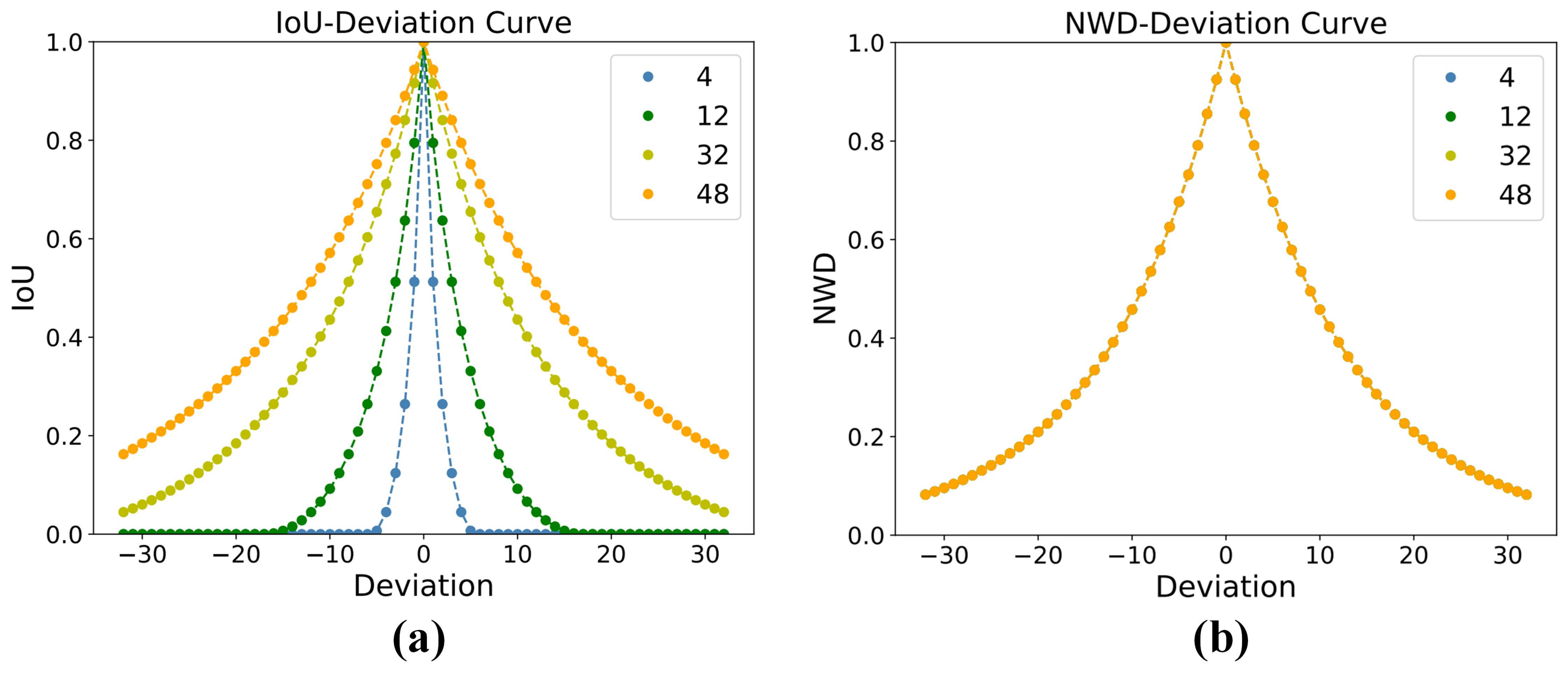}
	\end{center}
	\vspace{-5mm}
	\caption{A comparison between (a) the IoU-Deviation Curve and (b) the NWD Deviation Curve \cite{NWD}. Please refer to \cite{NWD} for detail.}
	\label{fig:NWD}
\end{figure}
	\subsection{Improved Detection Metrics for Tiny Object}
	Unlike the first two types of methods, recent advanced works \cite{R2CNN-TGRS, FSAN, YOLO-Fine, IoU-Adaptive,Sig-NMS, Center-Point-Guided, DoD, NWD, RFLA} assert that the current prevailing detection paradigms are unsuitable for tiny object detection and inevitably hinder tiny object detection performance. Pang \textit{et al.} \cite{R2CNN-TGRS} argued that the excessive down-sampling operations in modern detectors leads to the loss of tiny objects on the feature map and proposed a zoom-out and zoom-in structure to enlarge the feature map. 
	In \cite{IoU-Adaptive}, Yan \textit{et al.} adjusted the IoU threshold in the label assignment to increase the positive assigned anchors for tiny objects, facilitating the learning of tiny objects.  Dong \textit{et al.} \cite{Sig-NMS}  devised the Sig-NMS to reduce the suppression of tiny objects by large and medium objects in traditional non-maximum suppression (NMS).

	In \cite{NWD}, Xu \textit{et al.} pointed out that the IoU metric is unsuitable for tiny object detection. As shown in Fig. \ref{fig:NWD}, the IoU metric is sensitive to slight location offsets. 
	Besides, the IoU-based label assignment suffers from a severe scale imbalance problem, where tiny objects tend to be assigned with insufficient positive samples. 
	To solve these problems, Xu \textit{et al.} \cite{NWD} designed a normalized Wasserstein distance (NWD) to replace the IoU metric.  The NWD models the tiny objects as 2D Gaussian distributions and utilizes the normalized Wasserstein distance between Gaussian distributions to represent the location relationship between tiny objects, detailed in \cite{NWD}. 
	Compared with the IoU metric, the proposed NWD metric is smooth to location deviations and has the characteristic of scale balance, as depicted in Fig. \ref{fig:NWD}(b). In \cite{RFLA}, Xu \textit{et al.} further proposed the receptive field distance (RFLA) for tiny object detection and achieved state-of-the-art performance.
	\section{Object Detection with Limited Supervision}
\begin{figure}[t!]
		\begin{center}
			\includegraphics[width=0.47\textwidth]{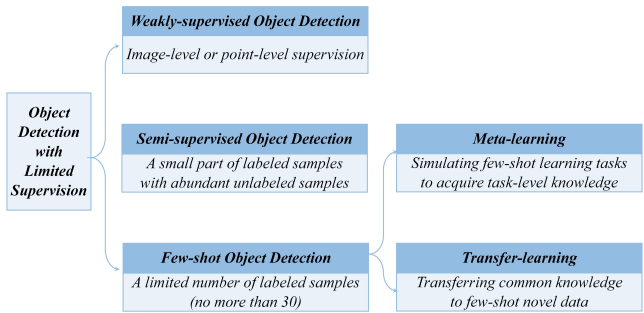}
		\end{center}
		\vspace{-3mm}
		\caption{A brief summary of object detection methods with limited supervision.}
		\label{fig:Data-limited-OD}
	\end{figure}

	In recent years, the widely used deep learning based detectors in RSIs heavily rely on large-scale datasets with high-quality annotations to achieve state-of-the-art performance. However, collecting volumes of well-labeled data is considerably expensive and time-consuming (e.g., a bounding box annotation would cost about 10 seconds), which leads to a data-limited or annotation-limited scenario in RSOD \cite{weak-supervision-survey}. This lack of sufficient supervised information seriously degrades the detection performance. To tackle this problem, researchers have explored various tasks in RSOD with limited supervision. We summarize the previous research into three main types:  weakly-supervised object detection, semi-supervised object detection, and few-shot object detection. Fig. \ref{fig:Data-limited-OD} provides a brief summary of object detection methods with limited supervision.

	\subsection{Weakly-supervised Object detection}

	Compared to fully supervised object detection, weakly-supervised object detection (WSOD) only contains weakly supervised information. Formally, WSOD consists of a training data set $\mathcal{D}_{\text {train}}=\left\{\left(X_{i}, y_{i}\right)\right\}_{i=1}^{I}$ , where $X_{i}=\left\{x_{1},...,x_{m_{i}} \right\}$ is a collection of training samples, termed as bag, $m_{i}$ is the total number of training samples in the bag, and $y_{i}$ is the weakly supervised information (e.g., image-level labels\cite{Coupled-CNN} or point-level labels \cite{Point-level}) of $X_{i}$. Effectively transferring image-level supervision to object-level labels is the key challenge in WSOD \cite{WSOD}.

	Han \textit{et al.} \cite{High-Level-Han} introduced a deep Boltzmann machine to learn the high-level features of objects and proposed a weakly-supervised learning framework based on Bayesian principles for remote sensing WSOD. Li \textit{et al.} \cite{ISPRS-Liyansheng} exploited the mutual information between scene pairs to learn the discriminative convolutional weights and employed a multi-scale category activation map to locate geospatial objects. 

\begin{figure}[t!]
	\begin{center}
		\includegraphics[width=0.46\textwidth]{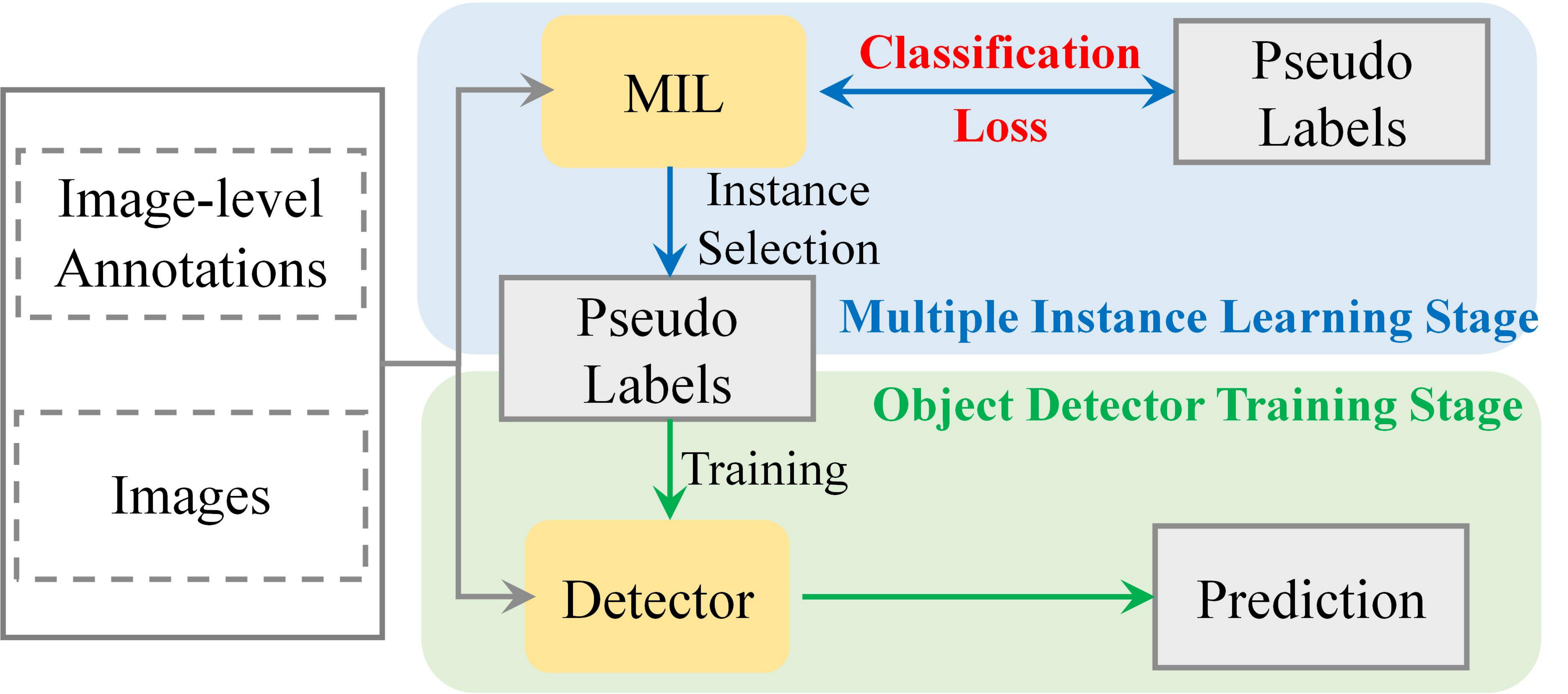}
	\end{center}
	\vspace{-3mm}
	\caption{The two-step paradigm of recent WSOD methods\cite{Automatic-Yaoxiwen, RS-Weakly2, Progressive-Fengxiaoxu, Multi-patch,Graph-Learning-Lixulong, TCANet, Self-supervised-Fengxiaoxu,  JSTAR-Weakly1, JSTAR-Weakly2, RS-Weakly1, Self-Guided,  Weakly-CVPR, MOL}.}
	\label{fig:WSOD}
\end{figure}

	Motivated by the remarkable performance of WSDDN \cite{WSDDN}, a series of remote sensing WSOD methods \cite{Automatic-Yaoxiwen, RS-Weakly2, Progressive-Fengxiaoxu, Multi-patch,Graph-Learning-Lixulong, TCANet, Self-supervised-Fengxiaoxu,  JSTAR-Weakly1, JSTAR-Weakly2, RS-Weakly1, Self-Guided,  Weakly-CVPR, MOL} are proposed. As shown in Fig. \ref{fig:WSOD}, the paradigm of the current WSOD methods usually consists of two steps, which first constructs a multiple instance learning model (MIL) to find contributing proposals to the image classification task as pseudo-labels and then utilizes them to train the detector. Yao \textit{et al.} \cite{Automatic-Yaoxiwen} introduced a dynamic curriculum learning strategy where the detector progressively improves detection performance through an easy-to-hard training process. Feng \textit{et al.} \cite{Progressive-Fengxiaoxu} designed a progressive contextual instance refinement method that suppresses low-quality object parts and highlights the whole object by leveraging surrounding context information. Wang \textit{et al.} \cite{Graph-Learning-Lixulong} introduced the spatial and appearance relation graph into WSOD, which propagates high-quality label information to mine more possible objects. In \cite{Weakly-CVPR}, Feng \textit{et al.} argued that existing remote sensing WSOD methods ignored the arbitrary orientations of geospatial objects, resulting in rotation-sensitive object detectors. To address this problem, Feng \textit{et al.} \cite{Weakly-CVPR} proposed a RINet, which brings rotation-invariant yet diverse feature learning for WSOD by employing rotation-invariant learning and multi-instance mining.

	We summarize the performance of milestone WSOD methods in Table \ref{summary-WSOD-results}, where the correct localization metric (CorLoc)\cite{CorLoc} is adopted to evaluate the localization performance.
	\begin{table}[t!]
	\renewcommand\arraystretch{1.4}
	\setlength\tabcolsep{4pt} 
	\footnotesize
	\centering		
	\caption{Performance of weakly supervised object detection methods on the NWPU VHR-10.v2 and DIOR datasets.}
	\vspace{-3mm}
	\begin{tabular}{ccccc}
		\hline
		\multirow{2}*{Models} &  \multicolumn{2}{c}{NWPU VHR-10}  &  \multicolumn{2}{c}{DIOR}   \\   
		&  CorLoc(\%) & mAP(\%)  &  CorLoc(\%) & mAP(\%) \\     
		\hline
		WSDDN \cite{WSDDN} &35.24 & 35.12 & 32.44 & 13.26\\
		DCL \cite{Automatic-Yaoxiwen} &69.65 & 52.11 &42.23 & 20.19\\
		DPLG \cite{RS-Weakly2} &61.50 & 53.60&-&-\\
		PCIR \cite{Progressive-Fengxiaoxu} & 71.87 & 54.97 &46.12 & 24.92\\
		MIGL \cite{Graph-Learning-Lixulong} & 70.16 &55.95 &46.80 &25.11\\
		TCANet \cite{TCANet} & 72.76 &58.82& 48.41 & 25.82 \\
		SAENet \cite{Self-supervised-Fengxiaoxu} &73.46 &60.72 & 49.42 & 27.10\\
		OS-DES \cite{JSTAR-Weakly1}&73.68 & 61.49&49.92 & 27.52\\
		SPG+MELM \cite{Self-Guided}  &73.41 & 62.80 & 48.30 & 25.77 \\
		RINet \cite{Weakly-CVPR} & -&70.4 & 52.8 & 28.3\\
		MOL \cite{MOL} & 75.96 & 75.46 & 50.66 & 29.21\\
		\hline
	\end{tabular}
	\vspace{-3mm}
	\label{summary-WSOD-results}
\end{table}

\subsection{Semi-supervised Object detection}
	Semi-supervised Object detection (SSOD) typically contains only a small part (no more than 50\%) of well-labeled samples $\mathcal{D}_{\text {labeled}}=\left\{\left(x_{i}, y_{i}\right)\right\}_{i=1}^{I_{labeled}}$, difficult to construct a reliable supervised detector, and has a large number of unlabeled samples $\mathcal{D}_{\text {unlabeled}}=\left\{\left(x_{j}\right)\right\}_{j=1}^{I_{unlabeled}}$. SSOD aims to improve detection performance under scarce supervised information by learning the latent information from volume unlabeled samples. 
	
\begin{figure}[t!]
	\begin{center}
		\includegraphics[width=0.46\textwidth]{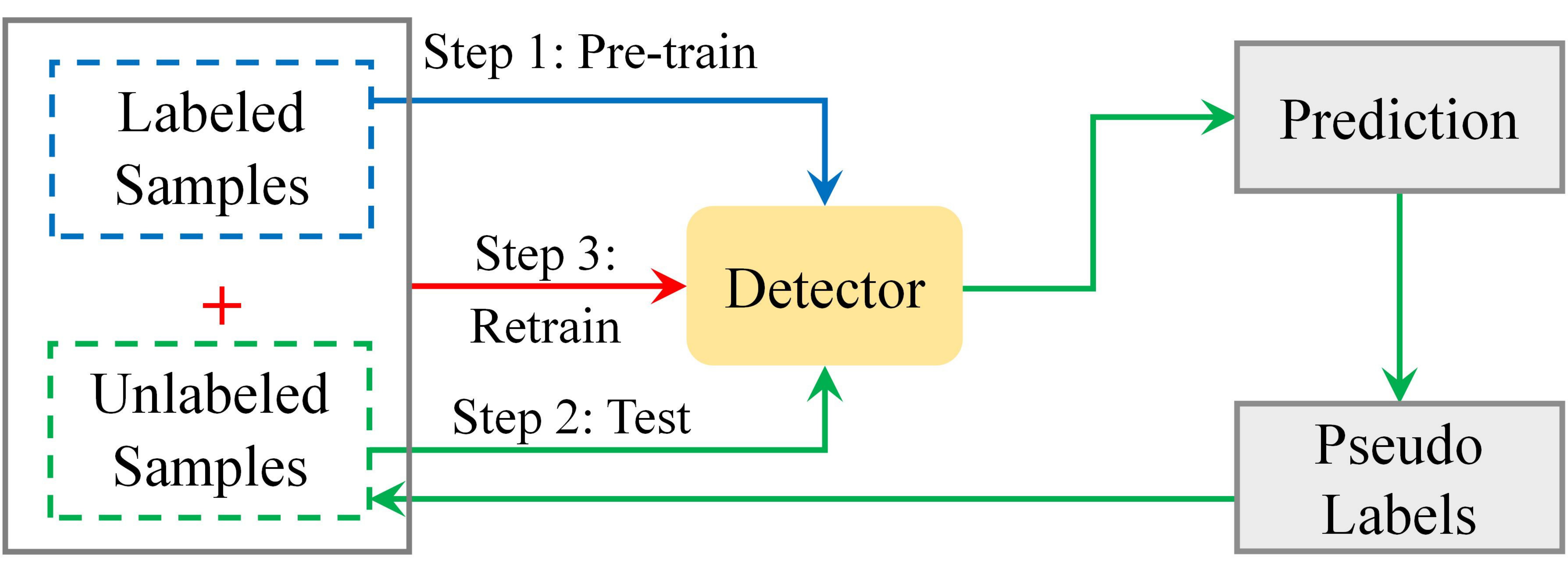}
	\end{center}
	\vspace{-3mm}
	\caption{The pipeline of pseudo-label generation mechanism in SSOD.}
	\label{fig:SSOD}
\end{figure}

	Hou \textit{et al.} \cite{Adversarial-Learning-Houbiao} proposed a SCLANet for semi-supervised SAR ship detection. The SCLANet employs adversarial learning between labeled and unlabeled samples to exploit the unlabeled sample information and adopts consistency learning for unlabeled samples to enhance the robustness of the network. The pseudo-label generation mechanism is also a widely used approach for semi-supervised object detection \cite{Pseudo-Label-Access, COLOR-Zhengzhuo, Curriculum-learning,Teacher-Student1, Teacher-Student2}, and the typical paradigm is shown in Fig. \ref{fig:SSOD}. First, a pre-trained detector learned from scare labeled samples are used to predict unlabeled samples, then the pseudo labels with higher confidence scores are selected as the trusted part, and finally, the model is retrained with the labeled and pseudo-labeled samples. Wu \textit{et al.} \cite{Curriculum-learning} proposed a self-paced curriculum learning that follows an “easy to hard” scheme to select more reliable pseudo labels. Zhong \textit{et al.} \cite{COLOR-Zhengzhuo} adopt an active learning strategy in which high-scored predictions are manually adjusted by experts to obtain refined pseudo labels. Chen \textit{et al.} \cite{Teacher-Student1} employed teacher-student mutual learning to fully leverage unlabeled samples and iteratively generate higher-quality pseudo-labels.
	
	In addition, some studies\cite{Hybird-CYB,Hybrid-SAR-Dulan1,Hybrid-SAR-Dulan2,Hybrid-SAR-Dulan3, Hybird-Instance-Segmentation} have worked on weakly semi-supervised object detection, in which the unlabeled samples are replaced with weakly annotated samples. Du \textit{et al.} \cite{Hybrid-SAR-Dulan2,Hybrid-SAR-Dulan3} employed a large number of image-level labeled samples to improve SAR vehicle detection performance under scarce box-level labeled samples. Chen \textit{et al.} \cite{Hybird-Instance-Segmentation} adopted a small portion of pixel-level labeled samples and a dominant amount of box-level labeled samples to boost the performance in label-scarce instance segmentation.

	\subsection{Few-shot Object Detection}
\begin{figure}[t!]
	\begin{center}
		\includegraphics[width=0.46\textwidth]{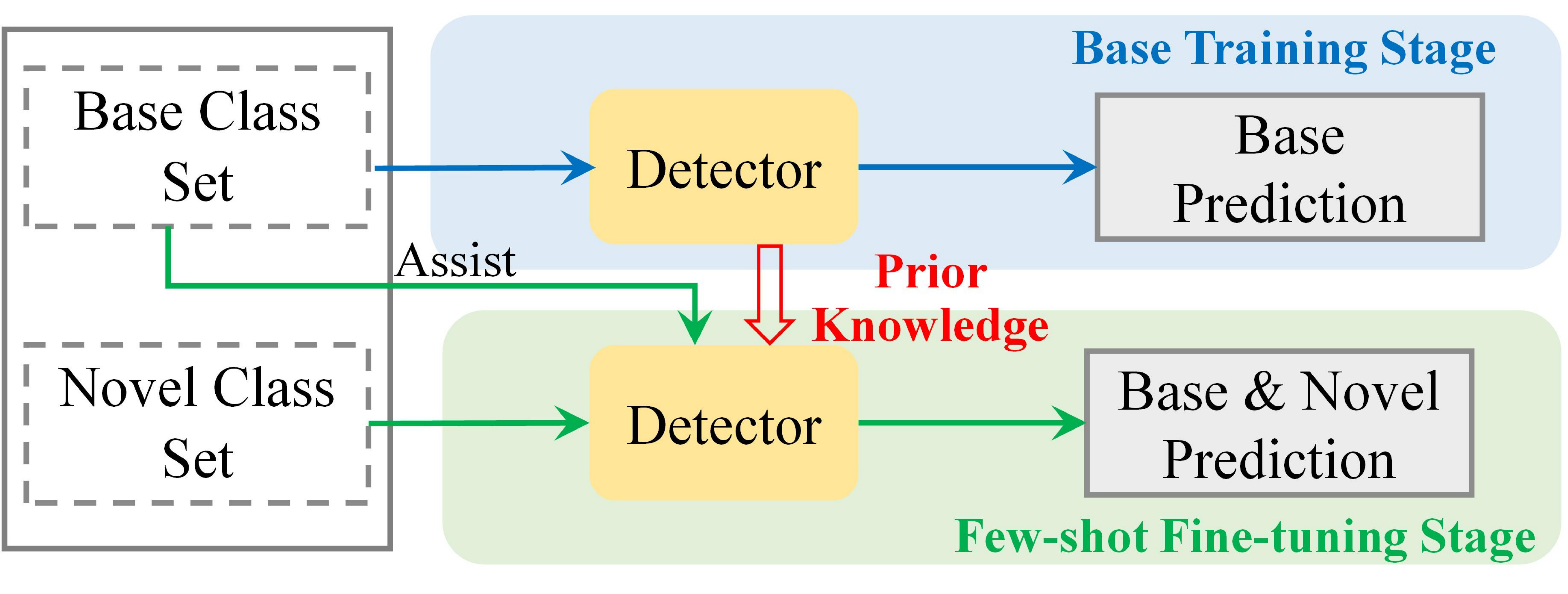}
	\end{center}
	\vspace{-3mm}
	\caption{The two-stage training pipeline of FSOD.}
	\label{fig:FSOD}
\end{figure}

	Few-shot object detection (FSOD) refers to detecting novel classes with only a limited number (no more than 30) of samples. Generally, FSOD contains a base class dataset with abundant samples $\mathcal{D}_{\text {base}}=\left\{\left(x_{i}, y_{i}\right), y_{i} \in C_{base} \right\}_{i=1}^{I_{base}}$ and a novel class dataset with only $K$-shot samples $\mathcal{D}_{\text {novel}}=\left\{\left(x_{j}, y_{j}\right), y_{j} \in C_{novel} \right\}_{j=1}^{C_{novel}*K}$. Note that $C_{base}$ and $C_{novel}$ are disjointed. As depicted in Fig. \ref{fig:FSOD}, a typical FSOD paradigm consists of a two-stage training pipeline where the base training stage establishes prior knowledge with abundant base class samples, and the few-shot fine-tuning stage leverages the prior knowledge to facilitate the learning of few-shot novel concepts. The research on remote sensing FSOD mainly focuses on meta-learning methods\cite{ Prototype-CNN, RS-FSOD,Oneshot,FSOD-meta-RS1, RS-FSIS, MM-RCNN} and transfer-learning methods\cite{FSOD-transfer-GRSL2, FSOD-transfer-GRSL1, FSOD-transfer-RS1, FSOD-transfer-RS2, FSOD-transfer-RS3,FSOD-transfer-RS4, FSOD-transfer-RS5,FSOD-transfer-JSTAR1,FSOD-ICCVW,G-FSDet}.
	
	The meta-learning based methods acquire task-level knowledge by simulating a series of few-shot learning tasks and generalize this knowledge to tackle few-shot learning of novel classes. Li \textit{et al.} \cite{RS-FSOD} first employed meta-learning for remote sensing FSOD and achieved satisfactory detection performance with only 1 to 10 labeled samples. Later, a series of meta-learning based few-shot detectors have been developed in the remote sensing community\cite{Prototype-CNN, RS-FSOD,Oneshot,FSOD-meta-RS1, RS-FSIS, MM-RCNN}. For example, Cheng \textit{et al.} \cite{Prototype-CNN} proposed a Prototype-CNN to generate better foreground proposals and class-aware RoI features for remote sensing FSOD by learning class-specific prototypes. Wang \textit{et al.} \cite{RS-FSIS} presented a meta-metric training paradigm to enable the few-shot learner with flexible scalability for fast adaptation to the few-shot novel tasks.
	
	Transfer-learning based methods aim at fine-tuning the common knowledge learned from the abundant annotated data to the few-shot novel data and typically consist of a base training stage and a few-shot fine-tuning stage. Huang \textit{et al.} \cite{FSOD-transfer-RS5} proposed a balanced fine-tuning strategy to alleviate the number imbalance problem between novel class samples and base class  samples. Zhou \textit{et al.} \cite{FSOD-transfer-RS4} introduced proposal-level contrast learning in the fine-tuning phase to learn more robust feature representations in few-shot scenarios. Compared with the meta-learning based methods, the transfer-learning based method has a simpler and memory-efficient training paradigm.

	\section{Datasets and Evaluation Metrics}
	\subsection{Datasets Introduction and Selection}
	Datasets have played an indispensable role throughout the development of object detection in RSIs. On the one hand, datasets serve as a common ground for the performance evaluation and comparison of detectors. On the other hand, datasets push researchers to address increasingly challenging problems in the RSOD field. In the past decade, several datasets with different attributes have been released to facilitate the development of RSOD, as shown in Table \ref{datasets}. In this section, we mainly introduce 10 widely used datasets with specific characteristics. 

	\begin{table*}[t!]
		\renewcommand\arraystretch{1.4}
		\small
		\centering
		\caption{Comparisons of widely used datasets in the field of RSOD. HBB and OBB refer to {\em horizontal bounding box} and {\em oriented bounding box}, respectively. $^{*}$ stands for {\em the average image width}.}
		\vspace{-3mm}
		\resizebox{1.0\linewidth}{!}{
			\begin{tabular}{c|ccccccccc}
				\hline
				Dataset &  Source & Annotation 
				& Categories
				
				& Instances & Images 
				& Image width &Resolution& Year                     
				\\ \hline
				
				TAS~\cite{TAS}                    
				& Google Earth    &HBB   & 1 & 1,319 
				&30 & 792  & - &  2008   \\
				
				SZTAKI-INRIA~\cite{SZTAKI}   
				
				&	Quick Bird, IKONOS and	Google Earth 
				
				& OBB &  1& 665 & 9  & $\sim$800  &0.5-1m &  2012  \\
				
				NWPU VHR-10~\cite{ML-1}             
				& Google Earth & HBB   & 10  & 3,651 
				& 800& $\sim$1,000     &0.3-2m & 2014  \\
				
				VEDAI~\cite{VEDAI}                 
				& Utah AGRC    & OBB    & 9 & 2,950 & 1,268    
				& 1,024  & 0.125m& 2015  \\
				
				DLR 3k~\cite{DLR3K}   
				& DLR 3K camera system  & OBB        &  8 & 14,235     
				& 20    & 5,616  & 0.13m  & 2015   \\
				
				UCAS-AOD~\cite{UCAS-AOD}   
				& Google Earth & OBB   & 2  & 6,029      & 910 & $\sim$1,000  &  0.3-2m  & 2015    \\
				
				COWC~\cite{COWC}                   
				&  Multiple Sources  & Point   & 1    & 32,716     
				& 53  & 2,000$-$19,000  &   0.15m   & 2016 \\
				
				HRSC~\cite{HRSC2016}           
				& Google Earth & OBB   & 26    & 2,976       
				& 1,061    & $\sim$1,100    & 0.4-2m & 2016  \\
				
				RSOD~\cite{RSOD}                  
				& Google Earth and Tianditu & HBB  & 4  & 6,950 & 976 
				& $\sim$1,000  & 0.3-3m  & \multicolumn{1}{l}{2017} \\
				
				SSDD~\cite{SSDD}                   
				& RadarSat-2, TerraSARX and Sentinel-1 & HBB    & 1   & 2,456     & 1,160   
				& 500    & 1-15m& \multicolumn{1}{l}{2017} \\	
				
				LEVIR~\cite{LEVIR} 
				& Google Earth & HBB   & 3  & 11,000 & 22,000   
				& 800$-$600  & 0.2-1m   & \multicolumn{1}{l}{2018} \\
				
				xView~\cite{xView}                 
				& Worldview-3  & HBB   & 60  & 1,000,000   & 1,413 
				& $\sim$3,000   & 0.3m  & \multicolumn{1}{l}{2018} \\
				
				DOTA-v1.0~\cite{DOTAv1}              
				& Google Earth, JL-1, and GF-2 & HBB and OBB  & 15& 188,282     & 2,806     & 800$-$13,000  &0.1-1m & 2018  \\
				
				HRRSD~\cite{Hierarchical-TGRS}                
				& Google Earth and Baidu Map & HBB & 13 & 55,740  & 21,761   
				& 152$-$10,569  & 0.15-1.2m   & 2019 \\
				
				DIOR~\cite{DIOR}                   
				& Google Earth & HBB    & 20   & 190,288     & 23,463   
				& 800    & 0.5-30m& \multicolumn{1}{l}{2019} \\
				
				AIR-SARShip-1.0~\cite{AirSARship}                   
				& Gaofen-3 & HBB    & 1   & 3,000     & 31   
				& 3,000   & 1m and 3m & \multicolumn{1}{l}{2019} \\			
					
				MAR20~\cite{MAR20}
				& Google Earth &	HBB and OBB   &	20 & 22,341 & 3,824 & $\sim$800 
				& - & \multicolumn{1}{l}{2020} \\
				
				FGSD~\cite{FGSD}
				& Google Earth &	OBB   &	43 & 5,634 & 2,612 & 930 
				& 0.12-1.93m & \multicolumn{1}{l}{2020} \\
				
				DOSR~\cite{DOSR}	   
				& Google Earth
				&	OBB   &	20 & 6,172 & 1,066 
				& 600-1,300  & 0.5-2.5m & \multicolumn{1}{l}{2021} \\	
				
				AI-TOD~\cite{AI-TOD}
				& Multiple Sources &HBB &	8 & 700,621 & 28,036 & 800 
				& - & \multicolumn{1}{l}{2021} \\
				
				FAIR1M~\cite{FAIR1M}	   
				& Gaofen satellites and Google Earth
				&	OBB   &	37 & 1,020,579 & 42,796 
				& 600-10,000 & 0.3-0.8m & \multicolumn{1}{l}{2021} \\	
				
				DOTA-v2.0~\cite{DOTAv2}                         
				& Google Earth, JL-1, GF-2 and airborne images & HBB and OBB    
				& 18  & 1,793,658   & 11,268   & 800-20,000 
				& 0.1-4.5m  & 2021     \\  
				
				SODA-A~\cite{SODA-A}
				&  Google Earth &	OBB   &	9 
				& 800,203 & 2,510 & 4,761$\times$2,777$^{*}$ & - & \multicolumn{1}{l}{2022} 			
				\\ \hline
			\end{tabular}
		}
		\vspace{-3mm}
		\label{datasets}
	\end{table*}

	\begin{table}[t!]
	\renewcommand\arraystretch{1.4}
	\setlength\tabcolsep{3.8pt} 
	\scriptsize
	\centering		
	\caption{Dataset selection guidelines in RSOD for different challenges and scenarios.}
	\vspace{-3mm}
	\begin{tabular}{ccc}
		\hline
		Scenarios  &  Datasets  &  Methods \\  
		\hline
		Multi-scale Objects   &DOTA, DIOR, FAIR1M & HyNet\cite{HyNet}, FFA\cite{ISPRS-czh}  \\
		Rotated Objects  & DOTA, HRSC & KLD\cite{RO20}, ReDet\cite{RI10} \\
		Weak Objects &DOTA, DIOR, FAIR1M &  RECNN\cite{RECNN}, CADNet\cite{CADNet} \\
		Tiny Objects  & SODA-A, AI-TOD & NWD\cite{NWD}, FSANet\cite{FSAN} \\
		Weakly Supervision  & NWPU VHR-10, DIOR & RINet\cite{Weakly-CVPR}, MOL\cite{MOL}\\
		Few-shot Supervision & NWPU VHR-10, DIOR & P-CNN\cite{Prototype-CNN}, G-FSDet\cite{G-FSDet}\\
		Fine-grained Objects  & DOSR, FAIR1M &  RBFPN\cite{Fine-grained-detection}, EIRNet \cite{DOSR}\\
		SAR image Objects  & SSDD, AIR-SARShip &  SSPNet\cite{SAR-Attention-TGRS-Sun},
		HyperLiNet\cite{HyperLi-Net} \\
		Specific Objects  & HRSC, MAR20 & GRS-Det\cite{RO23}, COLOR\cite{COLOR-Zhengzhuo} \\
		\hline
	\end{tabular}
	\vspace{-3mm}
	\label{Dataset-selection-guidelines}
\end{table}

\begin{figure*}[t!]
		\begin{center}
			\includegraphics[width=0.96\textwidth]{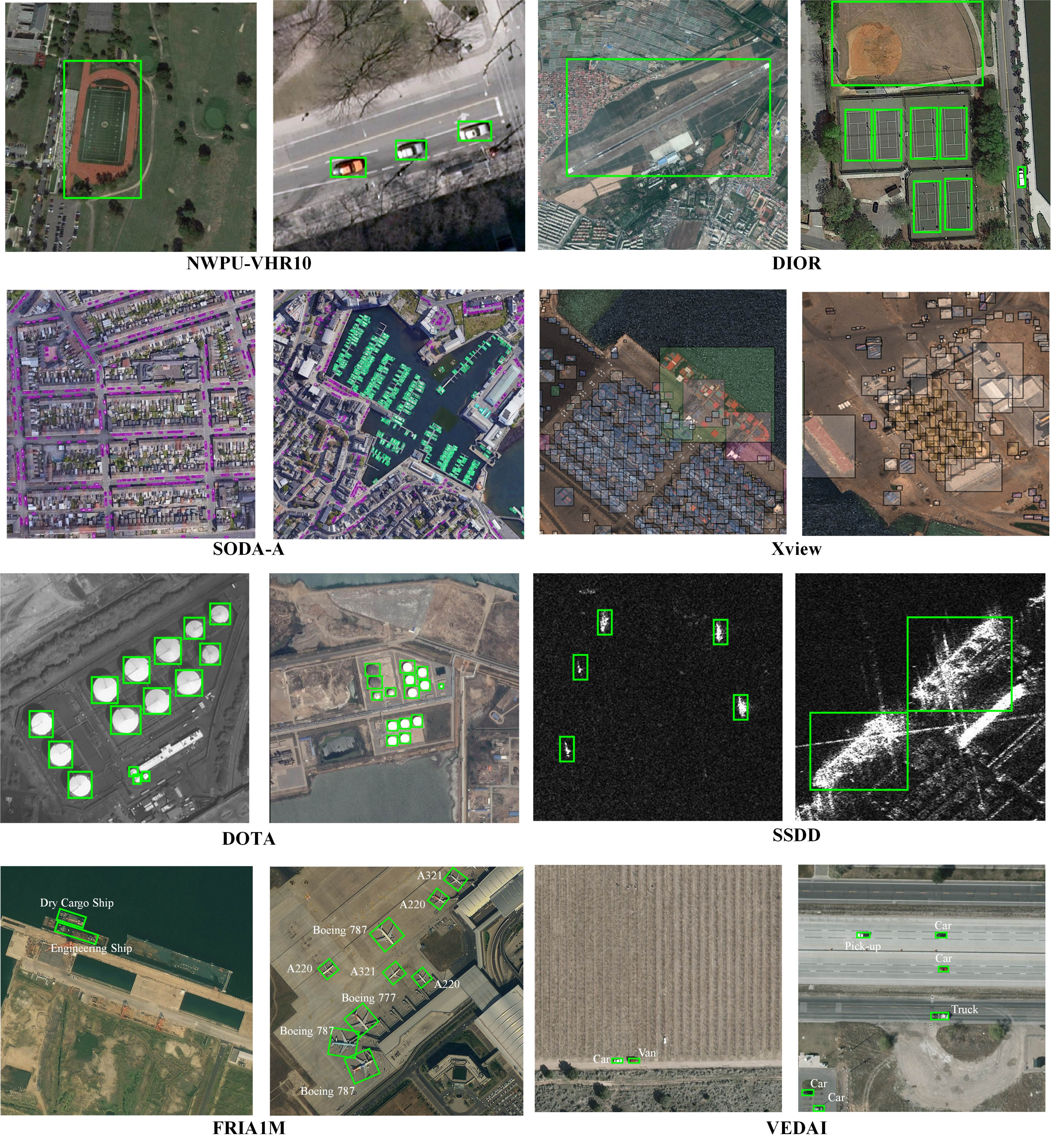}
		\end{center}
		\vspace{-3mm}
		\caption{Visualization of different RSOD datasets. Diverse resolutions, massive instances, multi-sensor images, and fine-grained categories are typical characteristics of RSOD datasets.}
		\label{fig:datasets}
\end{figure*}

\begin{figure*}[t!]
		\begin{center}
			\includegraphics[width=0.96\textwidth]{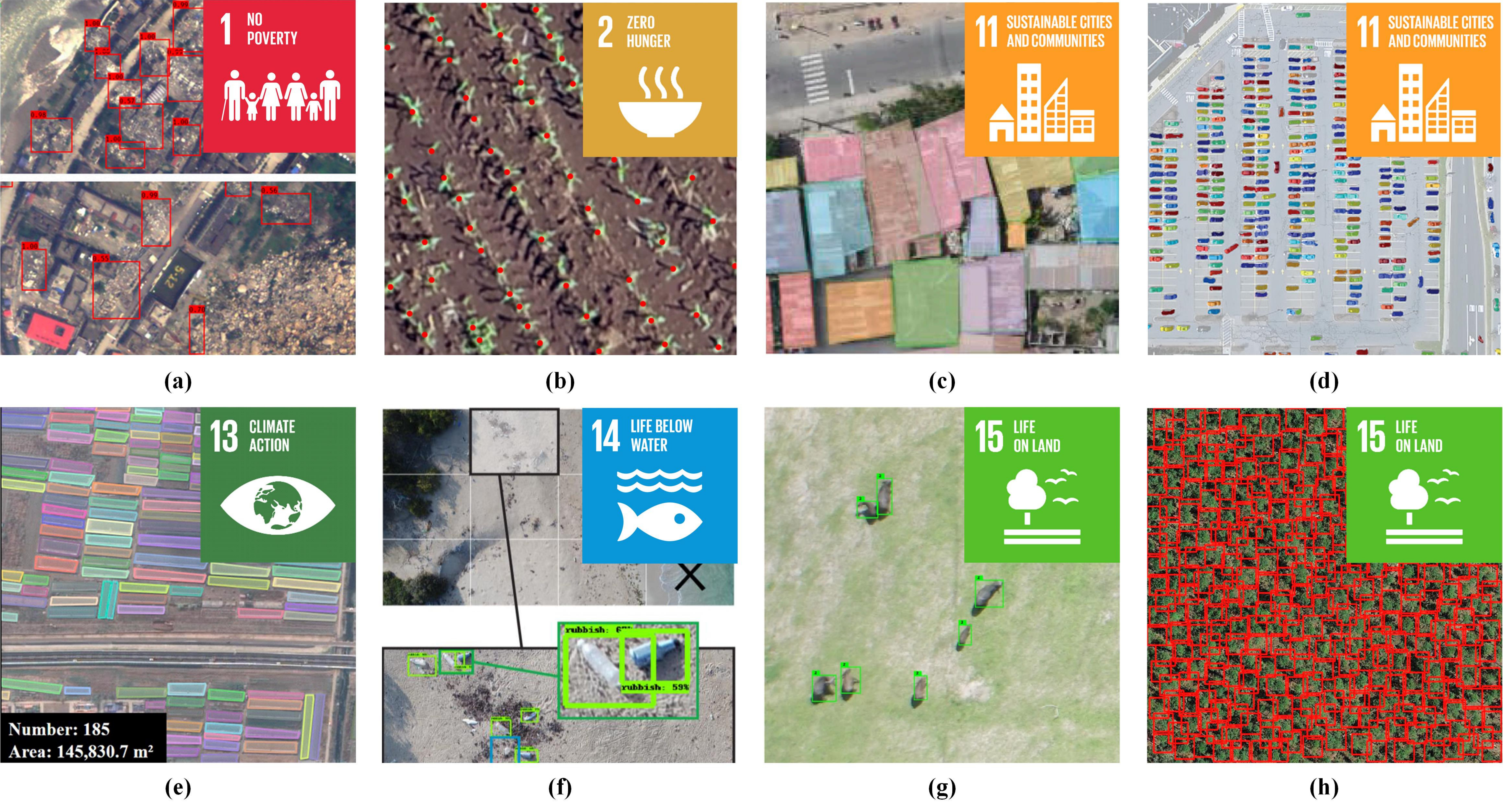}
		\end{center}
		\vspace{-3mm}
		\caption{The widespread applications of RSOD make substantial contributions to implementing of SDGs and improving society. 
		(a) Collapsed buildings detection in Post-Earthquake for disaster assessment.
		(b) Corn plant detection for precision agriculture.
		(c-d) Building and vehicle detection for sustainable cities and communities.
		(e) Solar photovoltaic detection for climate change mitigation. 
		(f) Litter detection along the shore for ocean conservation.
		(g) African mammals detection for wildlife surveillance.
		(h) Single tree detection for forest ecosystem protection.}
		\label{fig:applications}
\end{figure*}

	\textbf{NWPU VHR-10}\cite{ML-1}. This dataset is a multi-class geospatial object detection dataset. It contains 3,775 HBB annotated instances in ten categories: airplane, ship, storage tank, baseball diamond, tennis court, basketball court, ground track field, harbor, bridge, and vehicle. There are 800 very high-resolution RSIs, consisting of 715 color images from Google Earth and 85 pan-sharpened color infrared images from Vaihingen data. The image resolutions range from 0.5 to 2 m.
	
	\textbf{VEDAI}\cite{VEDAI}. VEDAI is a fine-grained vehicle detection dataset that contains five fine-grained vehicle categories: camping car, car, pick-up, tractor, truck, and van. There are 1,210 images and 3,700 instances in the VEDAI dataset, and the size of each image is $1,024 \times 1,024$. The small area and the arbitrary orientation of vehicles are the main challenges in the VEDAI dataset.

	\textbf{UCAS-AOD}\cite{UCAS-AOD}. The UCAS-AOD dataset includes 910 images and 6,029 objects, where 3,210 aircraft are contained in 600 images and 2,819 vehicles are contained in 310 images. All images are acquired from Google Earth with an image size of approximately $1,000 \times 1,000$.

	\textbf{HRSC}\cite{HRSC2016}. The HRSC dataset is widely used for arbitrary orientation ship detection and consists of 1,070 images and 2,976 instances with OBB annotation. The images are captured from Google Earth, containing offshore and inshore scenes. The image sizes vary from $300 \times 300$ and $1,500 \times 900$, and the image resolutions range from 2 to 0.4 m.

	\textbf{SSDD}\cite{SSDD}. SSDD is the first open dataset for SAR image ship detection and contains 1,160 SAR images and 2,456 ships. The SAR images in the SSDD dataset are collected from different sensors with resolutions from 1m to 15 m and have different polarizations (HH, VV, VH, and HV). Subsequently, the author further refines and enriches the SSDD dataset into three different types to satisfy the current research of SAR ship detection \cite{SSDD2}.
	
	\textbf{xView}\cite{xView}. The xView dataset is one of the largest publicly available datasets in ROSD, with approximately 1 million labeled objects across 60 fine-grained classes. Compared to other RSOD datasets, the images in xView dataset are collected from WorldView-3 at 0.3m ground sample distance, providing higher resolution images. Moreover, the xView dataset cover over 1,400 km$^2$ of the earth’s surface, which leads to higher diversity.
	
	\textbf{DOTA}\cite{DOTAv1}. DOTA is a large-scale dataset consisting of 188,282 objects annotated with both HBB and OBB. All objects are divided into 15 categories: plane, ship, storage tank, baseball diamond, tennis court, swimming pool, ground track field, harbor, bridge, large vehicle, small vehicle, helicopter, roundabout, soccer ball field, and basketball court. The images in this dataset are collected from Google Earth, JL-1 satellite, and GF-2 satellite with a spatial resolution of 0.1 to 1 m. Recently, the latest DOTAv2.0 \cite{DOTAv2} has been publicly available, which contains over 1.7 million objects of 18 categories.
	
	\textbf{DIOR}\cite{DIOR}. DIOR is an object detection dataset for optical RSIs. There are 23,463 optical images in this dataset with a spatial resolution of 0.5 to 30m. The total number of objects in the dataset is 192,472, and all objects are labeled with HBB. The categories of objects are as follows: airplane, airport, baseball field, basketball court, bridge, chimney, dam, expressway service area, expressway toll station, harbor, golf course, ground track field, overpass, ship, stadium, storage tank, tennis court, train station, vehicle, and windmill.

	\textbf{FAIR1M}\cite{FAIR1M}. FAIR1M is a more challenging dataset for fine-grained object detection in RSIs, including 5 categories and 37 subcategories. There are more than 40,000 images and more than 1 million objects annotated by oriented bounding boxes. The images are acquired from multiple platforms with a resolution of 0.3 m to 0.8 m and are spread across different countries and regions. The fine-grained categories, massive numbers of objects, large ranges of sizes and orientations, and diverse scenes make the FAIR1M more challenging.
	
	\textbf{SODA-A} \cite{SODA-A}. SODA-A is a recently released dataset designed for tiny object detection in RSIs. This dataset consists of 2,510 images with an average image size of $4,761 \times 2,777$, and 800,203 objects with OBB annotation. All objects are divided into four subsets (i.e., extremely small, relatively small, generally small, and normal) based on their area ranges. There are nine categories in this dataset, including airplane, helicopter, small-vehicle, large-vehicle, ship, container, storage-tank, swimming-pool, and windmill.
	
	The above review shows that the early published datasets generally have limited samples. For example, NWPU VHR-10\cite{ML-1} only contains 10 categories and 3,651 instances, and UCAC-AOD\cite{UCAS-AOD} consists of 2 categories with 6,029 instances. In recent years, researchers have not only introduced massive amounts of data and fine-grained level objects but also collected data from multi-sensor, various resolutions, and diverse scenes (e.g., DOTA\cite{DOTAv1}, DIOR\cite{DIOR}, FAIR1M\cite{FAIR1M}) to satisfy the practical applications in RSOD. Fig. \ref{fig:datasets} depicts the typical samples of different RSOD datasets.
	
	We also provide the dataset selection guidelines in Table \ref{Dataset-selection-guidelines} to help researchers select proper datasets and methods for different challenges and scenarios. Notably, only the image-level annotations of the datasets are available for the weakly supervision scenario. As for the few-shot supervision scenario, there are only $K$-shot box-level annotated samples for each novel class, where $K$ is set to $\left\{3,5,10,20,30  \right\}$.

	\subsection{Evaluation Metrics}
	In addition to the dataset, the evaluation metrics are equally important. Generally, the inference speed and the detection accuracy are the two commonly adopted metrics for evaluating the performance of detectors.
	
	\textbf{Frames Per Second} (FPS) is a standard metric for inference speed evaluation that indicates the number of images that the detector can detect per second. Notably, both the image size and hardware devices can influence the inference speed. 
	
    \textbf{Average Precision} (AP) is the most commonly used metric for detection accuracy. Given a test image $I$, let $\left\{\left(b_{i}, c_{i}, p_{i}\right)\right\}_{i=1}^{N}$ denotes the prediction detections, where $b_{i}$ is the predicted box, $c_{i}$ is the predicted label, and $p_{i}$ is the confidence score. Let $\left\{\left(b^{gt}_{j}, c^{gt}_{j}\right)\right\}_{j=1}^{M}$ refers to the ground truth annotations on the test image $I$, where $b^{gt}_{j}$ is the ground-truth box, and $c^{gt}_{j}$ is the ground truth category. A prediction detection $\left(b_{i}, c_{i}, p_{i}\right)$ is assigned as a True Positive (TP) for ground truth annotation $\left(b^{gt}_{j}, c^{gt}_{j}\right)$, if it meets both of the following criteria:
	\begin{itemize}
		\item The confidence score $p_{i}$ is greater than the confidence threshold $t$, and the predicted label is same as the ground truth label $c^{gt}_{j}$.
		
		\item The IoU between the predicted box $b_{i}$ and the ground truth box $b^{gt}_{j}$ is larger than the IoU threshold $\varepsilon$. The IoU is calculated as follows:

		\begin{equation}
		\mathrm{IoU}\left(b, b^g\right)=\frac{\operatorname{area}\left(b \cap b^g\right)}{\operatorname{area}\left(b \cup b^g\right)}
		\end{equation}
		
		where  $area(b_{i} \cap b^{gt}_{j})$ and $area(b_{i} \cup b^{gt}_{j})$ stand for the intersection and union area of the predicted box and ground truth box. 
	\end{itemize}

	Otherwise, it is considered to be a False Positive (FP). It is worth noting that multiple prediction detections may match the same ground truth annotation according to the above criteria, but only the prediction detection with the highest confidence score is assigned as a TP, and the rest are FPs\cite{PascalVOC}.
	
	Based on TP and FP detections, the Precision (P) and Recall (R) can be computed as Eq. \ref{Precision} and Eq. \ref{Recall}. 
	\begin{equation}
	P=\frac{TP}{TP+FP}
	\label{Precision}
	\end{equation}
	\begin{equation}
	R=\frac{TP}{TP+FN}
	\label{Recall}
	\end{equation}
	where \textit{FN} denotes the number of false negatives. The precision measures the fraction of true positives of the prediction detections and the recall  measures the fraction of positives that are correctly detected. However, the above two evaluation metrics only reflect the single aspect of detection performance. 

	Taking into account both precision and recall, AP provides a comprehensive evaluation of detection performance and is calculated individually for each class. For a given class, the Precision/Recall Curve (PRC) is drawn according to the detection of maximum Precision at each Recall, and the AP summarises the shape of the PRC \cite{PascalVOC}. For multi-class object detection, the mean of the AP values for all classes, termed $mAP$, is adopted to evaluate the overall detection accuracy.
	
	The early studies mainly employ a fixed IoU based AP metric (i.e., $AP_{50}$) \cite{ML-1, DIOR, DOTAv1}, where the IoU threshold $\varepsilon$ is given as 0.5. This low IoU threshold exhibits a high tolerance for bounding box deviations and fails to satisfy the high localization accuracy requirements. Later, some works \cite{RO19, RO20,SODA-A} introduce a novel evaluation metric, named $AP_{50:95}$, which averages the AP over 10 IoU thresholds from 0.5 to 0.95 with an interval of 0.05. The $AP_{50:95}$ considers higher IoU thresholds and encourages more accurate localization.
	
	As the cornerstone of evaluation metrics in RSOD, AP has various extensions for different specific tasks. In the few-shot learning scenario, $AP_{novel}$ and $AP_{base}$ are two critical metrics to evaluate the performance of few-shot detectors, where $AP_{novel}$ and $AP_{base}$ represent detection performance on the novel class and base class, respectively. An excellent few-shot detector should achieve satisfactory performance in the novel class and avoid performance degradation in the base class \cite{G-FSDet}. In the incremental detection of remote sensing objects, $AP_{old}$ and $AP_{inc}$ are employed to evaluate the performance of the old and incremental classes on different incremental tasks. In addition, the harmonic mean is also a vital evaluation metric for incremental object detection\cite{COD}, which provides a comprehensive performance evaluation of both old and incremental classes, as described by Eq. \ref{eq.HM}:
	\begin{equation}
	HM =\frac{2 A P_{old} AP_{inc}} {AP_{old} + AP_{inc}}
	\label{eq.HM}
	\end{equation}

	\section{Applications}
	Deep learning techniques have injected significant innovations into RSOD, leading to an effective way to automatically identify objects of interest from voluminous RSIs.
	Therefore, RSOD methods have been applied in a rich diversity of practice scenarios that significantly support the implementation of Sustainable Development Goals (SDGs) and the improvement of society\cite{applications-1, applications-2, applications-3}, as depicted in Fig. \ref{fig:applications}.

	\subsection{Disaster Management}
	Natural disasters pose a serious threat to the safety of human life and property. A quick and precise understanding of disaster impact and extent of damage is critical to disaster management.
	RSOD methods can accurately identify ground objects from a bird’s-eye view of the disaster-affected area, providing a novel potential for disaster management
	\cite{Review-Fire,Instance-Segmentation-Fire,Building-Damaged,Building-Damaged-YOLOv3,Disaster-Manage}. Guan \textit{et al}. \cite{Instance-Segmentation-Fire} proposed a novel instance segmentation model to accurately detect fire in a complex environment, which can be applied to the forest fire disaster response. Ma \textit{et al}. \cite{Building-Damaged-YOLOv3} designed a real-time detection method for collapsed building assessment in Post-Earthquake.
	
	\subsection{Precision Agriculture}
	With the unprecedented and still-expanding population, ensuring agricultural production is a fundamental obstacle to feeding the growing population. RSOD has the ability to monitor crop  growth and estimate food production, promoting further progress for precision agriculture \cite{Review-Precision-Agriculture, Crop-Pang, Corn-RS,Corn-ISPRS, Paddy-RS, Strawberry}. Pang \textit{et al.} \cite{Crop-Pang} used RSIs for early-season maize detection and achieved an accurate estimation of emergence rates. Chen \textit{et al.} \cite{Strawberry} designed an automatic strawberry flower detection system to monitor the growth cycle of strawberry fields.
	
	\subsection{Sustainable Cities and Communities}
	Half of the global population now lives in cities, and this population will keep growing in the coming decades.  Sustainable cities and communities are the goals of modern city development, in which RSOD can make a significant impact.  For instance, building and vehicle detection \cite{city-building-1, city-building-2, city-vehicle-1, city-vehicle-2} can help estimate population density distribution and transport traffic statistics, providing suggestions for city development planning. Infrastructure distribution detection \cite{city-disaster} can assist in  disaster assessment and early warning in the city environment.
	
	\subsection{Climate Action}
	The ongoing climate change forces humans to face the daunting challenge of the climate crisis. 
	Some researchers \cite{ice-1, ice-2, ice-3} employed object detection methods for automatically mapping tundra ice-wedge polygon to document and analyze the effects of climate warming on the Arctic region. Besides, RSOD can produce statistics on the number and spatial distribution of solar panels and wind turbines \cite{deepsolar,solar-2,wind-turbine-1, wind-turbine-2}, facilitating the mitigation of greenhouse gas emissions.
	\subsection{Ocean Conservation}
	The oceans cover nearly three-quarters of the Earth's surface, and more than 3 billion people depend on the diverse life of the oceans and coasts. The ocean is gradually deteriorating due to pollution, and the RSOD can provide powerful support for ocean conservation \cite{Review-Ocean}.
	Several works applied detection methods for litter detection along shores\cite{Shore}, floating plastic detection at sea \cite{Floating}, deep-sea debris detection\cite{Deep-Sea}, etc. Another important application is ship detection\cite{RO23, Ship-Tangxu}, which can help monitor illegal fishing activities.
	\subsection{Wildlife Surveillance}
	A global loss of biodiversity is observed at all levels, and object detection in combination with RSIs provides a novel perspective for wildlife conservation \cite{wildlife-1, wildlife-2, wildlife-3, wildlife-4, wildlife-5}. Delplanque \textit{et al.} \cite{wildlife-4} adopted the deep learning based detector for multi-species detection and identification of African mammals. Kellenberger \textit{et al.} \cite{wildlife-5} designed a weakly supervised wildlife detection framework that only requires image-level labels to identify wildlife.
	\subsection{Forest Ecosystem Protection}	
	The forest ecosystem plays an important role in ecological protection, climate regulation, and carbon cycling. Understanding the condition of trees is essential for forest ecosystem protection \cite{Review-Tree, Cascaded-Tree, Damaged-Tree, Mask-RCNN-Tree-1, Instance-Segmentation-Tree}. Safonova \textit{et al.} \cite{Damaged-Tree} analyzed the shape, texture, and color of the detected trees' crowns to determine their damage stage, providing a more efficient way to assess forest health. Sani-Mohammed \textit{et al.} \cite{Instance-Segmentation-Tree} utilized an instance segmentation approach to map the standing dead trees, which is imperative for forest ecosystem management and protection.

	\section{Future Directions}
	Apart from the five RSOD research topics mentioned in this survey, there is still much work to be done in this field. Therefore, we present a forward-looking discussion of future directions to further improve and enhance the detectors in remote sensing scenes.

	\subsection{Unified detection framework for large-scale remote sensing images}
	Benefiting from the development of remote sensing technology, high-resolution large-scale RSIs (e.g., over $10 ,000 \times 10,000$ pixels) can be easily obtained. However, limited by the GPU memory, the current mainstream RSOD methods fail to directly perform object detection in large-scale RSIs but adopt a sliding window strategy, mainly including sliding window cropping, patch prediction, and results merging. On the one hand, this sliding window framework requires complex data pre-processing and post-processing, compared with the unified detection framework. On the other hand, the objects usually occupy a small area of the RSIs, and the invalid calculation of the massive backgrounds leads to increasing computation time and memory consumption. Some studies \cite{R2CNN-TGRS, YOLT,CRNP-FSNET} proposed a coarse-to-fine detection framework for object detection in large-scale RSIs.
	This framework first locates the regions of interest by filtering out meaningless regions and then achieves accurate detection from these filtered regions.

	\subsection{Detection with Multi-modal remote sensing images}
	Restricted by the sensor imaging mechanism, the detectors based on the single-modal RSIs often have detection performance deviations, which are difficult to meet in practical applications \cite{MDMB-Hongdanfeng}. In contrast, the multi-modal RSIs from different sensors have their characteristics. For instance, hyperspectral images contain high spectral resolution and fine-grained spectral features, SAR images provide abundant texture information, and optical images exhibit high spatial resolution with rich detailed information. The integrated processing of multi-modal RSIs can improve the interpretation ability of the scene and obtain a more objective and comprehensive understanding of the geospatial objects \cite{ISPRS-Hongdanfeng, Multi-modal-RSE, multi-modal-ISPRS}, providing the possibility to further improve the detection performance of RSOD.
	\subsection{Domain adaptation object detection in remote sensing images}
	Due to the diversity of remote sensing satellite sensors, resolutions, and bands, as well as the influence of weather conditions, seasons, and geospatial regions \cite{zhu2017deep}, RSIs collected from different satellites are generally drawn from similar but not identical distributions. Such distribution differences (also called the domain gap) severely restrict the generalization performance of the detector. Recent studies on domain adaptation object detection \cite{DomainAdaptation-Dulan, DomainAdaptation-Fukun, DomainAdaptation-Sunxian, DomainAdaptation-RS} have proposed to tackle the domain gap problem. However, these studies only focus on the domain adaptation detectors in the single-modal, while the cross-modal domain adaptation object detection (e.g., from optical images to SAR images \cite{DomainAdaptation-SAR-Optical-Dulan, DomainAdaptation-SAR-Optical-JSTAR}) is a more challenging and worthwhile topic to investigate.
	\subsection{Incremental detection of remote sensing objects}
	The real-world environment is dynamic and open, where the number of categories evolved over time. However, mainstream detectors require both old and new data to retrain the model when meeting new categories, resulting in high computational costs. Recently, incremental learning has been considered the most promising way to solve this problem, which can learn new knowledge without forgetting old knowledge with only new data \cite{Incremental_1}. Incremental learning has been preliminarily explored in the remote sensing community \cite{Incremental_detection, Continual-RSOD, Incremental_2, Incremental_3}. For example, Chen \textit{et al.} \cite{Incremental_detection} integrated knowledge distillation into FPN and detection heads to learn new concepts while maintaining the old ones. More thorough research is still needed in incremental RSOD to meet the dynamic learning task in practical application.
	\subsection{Self-supervised pre-trained models for remote sensing scenes}
	Current RSOD methods are always initialized with the ImageNet \cite{ImageNet} pre-trained weights. However, there is an inevitable domain gap between the natural and remote sensing scenes, probably limiting the performance of RSOD. Recently, the self-supervised pre-training approaches have received extensive attention and shown excellent performance in the classification and downstream tasks in the nature scenes. Benefiting from the rapid advances in remote sensing technology, the abundant remote sensing data \cite{MillionAID, fMoW} also provide sufficient data support for self-supervised pre-training. Some researchers \cite{Self-supervised-Dubo, Geo-KR, Self-supervised-RingMo, Self-supervised-SatViT,Self-supervised-PVT} have initially demonstrated the effectiveness of remote sensing pre-training on representative downstream tasks. Therefore, exploring the self-supervised pre-training models based on multi-source remote sensing data deserves further research.
	\subsection{Compact and efficient object detection architectures}
	Most existing airborne and satellite-borne satellites require sending back the remote sensing data for interpretation, leading to additional resource overheads. Thus, it is essential to investigate compact and efficient detectors for airborne and satellite-borne platforms to reduce resource consumption in data transmission. Drawing on this demand, some researchers have proposed lightweight detectors through model design \cite{HyperLi-Net, ShipDetNet-20, Depthwise-Zhangtianwen}, network pruning \cite{Purning-Dulan, Purning-JSTAR}, and knowledge distillation \cite{Konwledge-Distillation-Fukun, Konwledge-Distillation-Sunxian, Instance-aware}. However, these detectors still rely heavily on high-performance GPUs and cannot be deployed on airborne and satellite-borne satellites. Therefore, designing compact and efficient object detection architectures for limited resources scenarios remains challenging.

	\section{Conclusion}
	Object detection has been a fundamental but challenging research topic in the remote sensing community. Thanks to the rapid development of deep learning techniques, RSOD has received considerable attention and gained remarkable achievements in the past decade. In this review, we present a systematic review and summarization of existing deep learning based methods in RSOD. Firstly, we summarized the five main challenges in RSOD according to the characteristics of geospatial objects and categorized the methods into five streams: multi-scale object detection, rotated object detection, weak object detection, tiny object detection, and object detection with limited supervision. Then, we adopted a systematic hierarchical division to review and summarize the methods in each category. 	
	Next, we introduced the typical benchmark datasets, evaluation metrics, and practical applications in the RSOD field.
	Finally, considering the limitations of existing RSOD methods, we discussed some promising directions for further research.
	
	Given this time of high-speed technical evolution in RSOD, we believe this survey can help researchers to achieve a more comprehensive understanding of the main topics in this field and to find potential directions for future research.

	\section*{Acknowledgement}
	This work was supported in part by the National Natural Science Foundation of China, under Grant 62276197, Grant 62006178, Grant 62171332; in part by the Key Research and Development Program in the Shaanxi Province of China under Grant 2019ZDLGY03-08.
	Xiangrong Zhang is the corresponding author.
	
	\ifCLASSOPTIONcaptionsoff
	\fi
	\bibliographystyle{IEEEtran}
	\bibliography{IEEEabrv,mybibfile}
	
\end{document}